\documentclass{article}

\usepackage{nicematrix}
\usepackage{multirow}

\usepackage{amsmath}  
\usepackage{bm}
\usepackage{wrapfig}
\usepackage{float}
\usepackage{algorithm}  
\usepackage{algpseudocode}

\usepackage{subcaption}

\usepackage[preprint]{neurips_2025}

\usepackage[utf8]{inputenc} 
\usepackage[T1]{fontenc}    
\usepackage{hyperref}       
\usepackage{url}            
\usepackage{booktabs}       
\usepackage{amsfonts}       
\usepackage{nicefrac}       
\usepackage{microtype}      
\usepackage{xcolor}         

\title{Towards Latent Diffusion Suitable For Text}

\author{%
  Nesta Midavaine \\
  University of Amsterdam\\
  \texttt{nesta.midavaine@student.uva.nl} \\
  \And
  Christian A. Naesseth \\
  University of Amsterdam \\
  \texttt{c.a.naesseth@uva.nl} \\
  \And
  Grigory Bartosh \\
  University of Amsterdam \\
  \texttt{g.bartosh@uva.nl} \\
}

\begin{document}

\maketitle

\begin{abstract} 
Language diffusion models aim to improve sampling speed and coherence over autoregressive LLMs. We introduce Neural Flow Diffusion Models for language generation, an extension of NFDM that enables the straightforward application of continuous diffusion models to discrete state spaces. NFDM learns a multivariate forward process from the data, ensuring that the forward process and generative trajectory are a good fit for language modeling. Our model substantially reduces the likelihood gap with autoregressive models of the same size, while achieving sample quality comparable to that of previous latent diffusion models. The code is available at \nolinkurl{https://github.com/Nesta-gitU/discrete_diffusion}.
\end{abstract}

\section{Introduction}
Language diffusion models go beyond next-token prediction by modeling text jointly, improving sampling speed for long sequences \cite{google_deepmind_2025} and overcoming the reversal curse \cite{nie_large_2025}. Existing methods achieve this by relying on predefined noise strategies \cite{campbell_continuous_2022, gat_discrete_2024} or on noise schedules guided by linguistic priors \cite{chen_cheaper_2023, shaul_flow_2024}. These methods require manual adjustments to the schedule to accommodate different data distributions. We propose learning the noising process directly from data, allowing the noise schedule to adapt conditionally to the input.

Language diffusion models fall into two families: discrete models that diffuse directly over tokens \cite{gat_discrete_2024,hoogeboom_autoregressive_2022,campbell_continuous_2022} and latent models that operate in continuous embedding space \cite{gulrajani_likelihood-based_nodate,li_diffusion-lm_2022,shabalin_tencdm_2025}. We introduce Neural Flow Diffusion Models (NFDM) for language generation, extending NFDM \cite{bartosh_neural_flow_2024} to language. Instead of a fixed global schedule, NFDM learns a multivariate, data-conditioned forward process, yielding per-token noise schedules and a theoretically lower achievable ELBO compared to other continuous latent space approaches \cite{bartosh_neural_2024,bartosh_neural_flow_2024,sahoo_diffusion_2024}. To demonstrate the effectiveness of our approach, we train several diffusion models with different learnable forward processes. We compare these to an autoregressive and static forward process baseline. We find that our best model significantly reduces the likelihood compared to the static forward process model and is within one standard deviation of the similarly sized autoregressive baseline. While we achieve sample quality comparable to previous diffusion model baselines.
\section{Background} \label{sec:related_work}



\paragraph{Score-based diffusion models} define a forward a Stochastic Differential Equation (SDE) that perturbs data $\bm{x} \sim q(\bm{x})$ into Gaussian noise through a trajectory of variables $\{\bm{z}(t)\}_{t\in[0,1]}$  \citep{song_score-based_2021}.
\begin{equation}
    d \bm{z}_t = f(\bm{z}_t,t)\,dt + g(t)\,d\bm{w}, \quad \bm{z}_0 \approx \bm{x},
\end{equation}
Here ${f}(z_t, t) : \mathbb{R}^D \times [0, 1] \to \mathbb{R}^D$ is a linear in $\bm{z}_t$ drift term,  
$g(t) : [0, 1] \to \mathbb{R}_{+}$ a scalar variance and $\bm{w}$ is a standard Wiener process. To generate samples, the SDE is reversed \citep{song_score-based_2021}. 
\begin{equation}
    d \bm{z}_t = \big(f(\bm{z}_t,t) - g^2(t)\,\nabla_{\bm{z}_t}\log q_t(\bm{z}_t)\big) dt + g(t)\,d\bar{\bm{w}}.
\end{equation}
Here, $\bar{\bm{w}}$ denotes a standard Wiener process where time flows backward. Since the score $\nabla_{\bm{z}_t}\log q_t(\bm{z}_t)$ is unknown, a neural network is trained via denoising score matching over noise scales.
\paragraph{Neural Flow Diffusion Models} NFDM extend Score-based diffusion models by allowing the forward process to be learned \citep{bartosh_neural_flow_2024}. 
The conditional forward SDE is given implicitly via a learnable transformation $F_\varphi(\varepsilon, t, \bm{x})$. 
More specifically, NFDM characterizes the marginal distributions $q_{\varphi}(\bm{z}_t|\bm{x})$ of the forward process through,
\begin{equation}
    \bm{z}_t = F_{\varphi}(\epsilon, t, \bm{x}), \quad \text{$\epsilon \sim q(\epsilon) = \mathcal{N}(\epsilon;0,I)$}
\end{equation}
During training, this formulation allows $\bm{z}_t \sim q_{\varphi}(\bm{z}_t|\bm{x})$ to be sampled efficiently. To completely define the distribution of the trajectories $\{\bm{z}(t)\}_{t \in [0,1]}$, NFDM introduces the forward conditional SDE,
\begin{align} \label{forwardsde}
    \mathrm{d} \bm{z}_t = \tilde{f}^{F}(\bm{z}_t,t, \bm{x})\,dt + g^{2}_{\varphi}(t)\,d \bm{w}, \quad \text{where} \\
    \tilde{f}^{F}(\bm{z}_t,t,\bm{x})=f(\bm{z}_t,t, \bm{x})+\frac{g^{2}_{\varphi}(t)}{2}\,\nabla_{\bm{z}_t}\log q_{\varphi}(\bm{z}_t|\bm{x}). \label{drift}
\end{align}
\autoref{drift} shows the forward drift term $\tilde{f}^{F}(z_t,t):\mathbb{R}^{D}\!\times[0,1]\!\to\!\mathbb{R}^{D}$, where $f(\bm{z}_t,t, \bm{x})$ and $\nabla_{\bm{z}_t}\log q_{\varphi}(\bm{z}_t|\bm{x})$ can be derived through $F_\varphi$.
NFDM additionally defines a learnable reverse generative process. This framework enables the model to adapt to the data distribution while maintaining a tractable and simulation-free training process.

\section{Method}

\subsection{NFDM in latent space} \label{sec:method subsec:nfdminlatentspace}
Neural Flow Diffusion Models define a learnable, continuous forward process dependent on the given data $\bm{x}$. In our case, $\bm{x}$ represents a sequence of discrete tokens. We parameterize a continuous forward process by using an embedding matrix $E_\varphi({\bm{x}})$ and defining $F_\varphi$ as a linear in $\epsilon$ transformation that progressively destroys information until it reaches a simple noise-like distribution at $t=1$. The functions $\mu$ and $\sigma$ in \autoref{latentdiffusionF} are parameterized as non-linear transformations of the input embeddings. Ensuring that the overall signal decays appropriately during the forward process, the exact parameterization of $F_\varphi$ used in NFDM is provided in Appendix \ref{app: nfdm exact}.

\begin{equation} \label{latentdiffusionF}
F^{\text{Gaussian}}_{\varphi}(\epsilon, t, E_\varphi({\bm{x}})) = \mu_{\varphi}(E_\varphi({\bm{x}}), t) + \sigma_{\varphi}(E_\varphi({\bm{x}}), t) \epsilon.
\end{equation}

\noindent The distribution of $\bm{z}_0$ at $t=0$ is centered on the embeddings with small variance. Since the embeddings are learned, we must include a reconstruction loss during training. Prior work \cite{gao_empowering_2024, dieleman_continuous_2022} finds that if this loss term is excluded, the embeddings can collapse to the same point. To get a generative model, we can parameterize the reverse flow $\tilde{f}_{\varphi}^{B}$ with the predicted embeddings $\hat{E}_\theta(\bm{z}_t, t)$ as done in \autoref{reverse_flow_hat}. We can then learn the parameters of this predictor jointly with the parameters of the forward process using the NFDM diffusion loss in \autoref{NFDM elbo latent2}.

\begin{equation} \label{reverse_flow_hat}
     \hat{f}_{\theta, \varphi}^{B}(\bm{z}_t, t) = f(\bm{z}_t,t, \hat{E}_\theta(\bm{z}_t, t))-\frac{g^{2}_{\varphi}(t)}{2}\,\nabla_{\bm{z}_t}\log q_{\varphi}(\bm{z}_t|\hat{E}_\theta(\bm{z}_t, t))
\end{equation}

\begin{align} \label{NFDM elbo latent2}
\tilde{\mathcal{L}}_{\text{diff}} = \mathbb{E}_{u(t) \, q(\bm{x}) \, q_{\varphi}(\bm{z}_t \mid \bm{x})} 
\Bigg[
    \frac{1}{2 g_{\varphi}^2(t)}
    \big\|
        \tilde{f}_{\varphi}^{B}(\bm{z}_t, t, \bm{x})) 
        - \hat{f}_{\theta, \varphi}^{B}(\bm{z}_t, t)
    \big\|_2^2
\Bigg]
\end{align}


In \citep{bartosh_neural_flow_2024}, it is shown that several existing approaches that learn the diffusion forward process can be viewed as special cases of NFDM. Among these, we experiment with MuLAN \citep{sahoo_diffusion_2024}, which defines a dimension-wise multiplicative Gaussian forward process. In this formulation, the mean and variance functions $(\mu, \sigma)$ in \autoref{latentdiffusionF} are linear and parameterized such that each token position in the sequence maintains its own noise parameters. Consequently, the model can learn a soft noising order that adapts to the input. When adopting the linear forward process of MuLAN, the volatility term $g^2(t)$ can be analytically derived, yielding a simplified diffusion loss:
\begin{align} \label{mulan loss}
\mathcal{L}^{\text{MuLAN}}_\text{diff} &= \lambda_F(t) \left\| E_\varphi({\bm{x}}) - \hat{E}_\theta({\bm{z}_t,t})\right\|_2^2 
\end{align}
\noindent Here $\lambda_F(t)$ is determined by the definition of $F_\varphi$, see Appendix~\ref{details of mulan} for a detailed derivation.
Previous diffusion language modeling works often employ a rescaled loss $\mathcal{L}_{\bm{x}} = \tfrac{1}{\lambda_F}\mathcal{L}^{\text{MuLAN}}_{\text{diff}}$ \cite{li_diffusion-lm_2022, shabalin_tencdm_2025}. However, in the general NFDM setting, such rescaling can cause training collapse, as the model will avoid early noise injection when this avoidance is not penalized. In contrast, MuLAN’s per-dimension parameterization enables the stabilization of training even under rescaled loss. We fix a global signal-to-noise ratio (SNR) schedule \cite{kingma_variational_2023} while allowing the model to learn how this global signal is distributed across dimensions. This design constrains the overall noise while enabling dimension-specific information retention. Implementation details for maintaining a fixed-average SNR under the rescaled loss are provided in Appendix~\ref{global SNR}.


\subsection{Sampling}
\begin{wrapfigure}{r}{0.5\textwidth}
    \centering
    \begin{minipage}{0.5\textwidth}
        \begin{algorithm}[H]
            \caption{DDIM-style sampling \cite{song_denoising_2022}}
            \label{discrete_sampling}
            \begin{algorithmic}
                \Require $F_\varphi$, $g_\varphi$, $\bm{x}$, $T$, $E_\varphi$
                \State $\Delta t = 1/T$, $\textbf{z}_1 \sim p(\textbf{z}_1)$
                \For{ $t=1, \ldots, \frac{2}{T}, \frac{1}{T}$}
                    \State $s = t - \Delta t$
                    \State $\epsilon = F_{\varphi}^{-1}(\bm{z}_t, t, E_\varphi({\bm{x}}))$
                    \State $\bm{z}_s = F_\varphi(\epsilon, s, \hat{E}_\theta(\bm{z}_t, t))$, $\bm{z}_t = \bm{z}_s$
                \EndFor
                \State $\bm{x} \sim p_\varphi(\bm{x}|\bm{z}_0;E_\varphi)$
            \end{algorithmic}
        \end{algorithm}
    \end{minipage}
\end{wrapfigure}
One way to generate a sequence from NFDM is to sample from the prior distribution and simulate the learned reverse SDE. Alternatively, we can sample from a sequence of marginal distributions that match those of the SDE, as outlined in Algorithm \ref{discrete_sampling}. This approach introduces several sources of bias. Each latent state is drawn from its marginal rather than conditioned on the previous state. Moreover, relying on a single model prediction, $\hat{E}_\theta(\bm{z}_t, t)$, instead of sampling from the full conditional distribution, pulls samples toward the center of the data distribution. These sampling choices reduce variance and bias sample generation toward high-probability regions. This can be advantageous when targeting low perplexity under a teacher model. Finally, once $\bm{z}_0$ is obtained, the decoder $p_\varphi(\bm{x}|\bm{z}_0; E_\varphi)$ is used to map the embeddings back to discrete tokens. In our setting, the decoder inverts the encoder by assigning each latent vector to its closest vocabulary embedding, measured via dot-product similarity.

\section{Experiments}
We compare the different parameterizations of the forward process discussed above to prior work based on likelihood and sample quality. Comparisons are conducted for the task of unconditional generation on the ROCstories dataset \cite{mostafazadeh_corpus_2016}, which comprises 90,000 examples of common-sense life stories.

\subsection{Experimental setup}

\paragraph{Models} For the model that predicts the embeddings used to parameterize the reverse dynamics, we use BERT-base \cite{devlin_bert_2019} with timestep embeddings, following the setup of \cite{li_diffusion-lm_2022}. To parameterize the transformation $F^\text{Gaussian}_\varphi$ in Latent-NFDM, we use a smaller BERT-style model with around 10 million parameters. This model is conditioned on time using Adaptive LayerNorm conditioning  \cite{peebles_scalable_2023, perez_film_2017}. More detailed modeling and hyperparameter choices for training are reported in Appendix \ref{implementation details}.
\paragraph{Evaluation} Following \cite{lovelace2023latent}, we report GPT-2 Large perplexity \cite{radford2019language}, MAUVE \cite{pillutla_mauve_2021}, diversity, memorization, and bits-per-character (ELBO). We follow \citep{li_diffusion-lm_2022} and use 2000 sampling steps. To compare likelihoods, we compute bits per character (bpc) in terms of ELBO.
\paragraph{Baselines} We compare Latent-NFDM and MuLAN against Diffusion-LM \citep{li_diffusion-lm_2022}, a diffusion-based language model that performs Gaussian diffusion directly in the embedding space. Diffusion-LM employs a static forward process and discrete timesteps in its original formulation. For consistency, we reformulate Diffusion-LM as a special case of NFDM and use this version for all training and evaluation, see Appendix~\ref{App: diff-LM} for details. We additionally evaluate samples from the ROCstories test set. 
For likelihood evaluation, we additionally report results for GPT-J and Diffusion-LM-Likelihood from \citep{li_diffusion-lm_2022}. GPT-J is an autoregressive transformer trained from scratch on the same dataset, with a comparable parameter count to our models. Diffusion-LM-Likelihood is a variant of Diffusion-LM specifically optimized for density estimation. 

\subsection{Experimental results} \label{maintextresults}
\autoref{likelihoodtable} shows the likelihood results. The standard MuLAN performs worse than the rescaled version, while MuLAN-rescaled and NFDM outperform both Diffusion-LM baselines. In doing so, NFDM reduces the likelihood gap towards the autoregressive GPT-J to almost zero. 

For sample quality, we examine \autoref{discrete sampling}, which presents the discrete sampling results for the different models. We observe that MuLAN-Rescaled and NFDM both closely match the test set in terms of perplexity, diversity, and memorization. Diffusion-LM performs better in terms of perplexity and MAUVE. However, it should be noted that it achieves lower perplexity than the test set, which indicates diffusion-LM might be overly biased towards the mean of the language distribution. In terms of sample quality, MuLAN-rescaled outperforms the version trained without the rescaled loss by a large margin. When we compare the results of NFDM to MuLAN, we observe that the additional flexibility of NFDM yields a considerable improvement in sample quality. When we examine diversity, we see that none of the models have scores that indicate repetitive samples. In Appendix \ref{additional results} we experiment with different sampling methods and hyperparameters. 

\begin{table}[H] 
\centering
\caption{Bits Per Character computed in terms of ELBO.}
\begin{tabular}{lccccc}
\toprule
Model & ELBO (SE) \\
\midrule
GPT-J* & 3.05 (-)\\
\midrule
Diffusion-LM-Likelihood* & 3.88 (-) \\
Diffusion-LM & 5.94 (0.41)  \\
MuLAN & 5.78 (0.07) \\
MuLAN-Rescaled & 3.47 (0.09) \\
NFDM & 3.12 (0.05)\\
\bottomrule
\end{tabular}
\label{likelihoodtable}
\end{table}

\begin{table}[H]
\centering
\caption{Sample quality results for DDIM-style sampling.}
\begin{tabular}{lccccc}
\toprule
Model & PPL $\downarrow$ & MAUVE $\uparrow$ & Diversity $\uparrow$ & Memorization $\downarrow$ \\
\midrule
Diffusion-LM & 21.05 (0.24) & 0.76 (0.01) & 0.18 (0.00) & 0.14 (0.00) \\
MuLAN & 139.74 (1.81) & 0.01 (0.00) & 0.21 (0.00) & 0.02 (0.00) \\
MuLAN-Rescaled & 27.71 (0.31) & 0.60 (0.03) & 0.20 (0.00) & 0.12 (0.00) \\
NFDM & 26.44 (0.22) &  0.43 (0.03) & 0.11 (0.00) & 0.14 (0.00) \\
\midrule
Dataset & 26.106  & 0.957 & 0.200 & 0.141 \\
\bottomrule
\end{tabular}
\label{discrete sampling}
\end{table}

\section{Related Work}
Diffusion models generate data by reversing a noise process \citep{ho_denoising_2020, song_score-based_2021}. Diffusion models can achieve strong likelihoods by learning the SNR of this process, as done in VDM \cite{kingma_variational_2023}. Latent diffusion is a straightforward method for extending diffusion models to language. Diffusion-LM embeds tokens and applies Gaussian diffusion with a square-root noise schedule \citep{li_diffusion-lm_2022}, CDCD \citep{dieleman_continuous_2022} trains with cross-entropy to encourage commitment to specific tokens. TEncDM diffuses in contextual language-model encodings and reports strong results \citep{shabalin_tencdm_2025}. Complementary to these latent approaches, recent advances in diffusion modeling, such as NFDM and MuLAN, learn multivariate, input-dependent forward processes with per-dimension/per-token noise allocation, thereby tightening ELBOs \citep{bartosh_neural_flow_2024, sahoo_diffusion_2024}.

\section{Conclusion}
We proposed Neural Flow Diffusion Models for language generation, extending NFDMs by learning the multivariate, data-conditioned forward process in latent space. Empirically, NFDM closes much of the likelihood gap with autoregressive models of comparable size while achieving comparable sample quality to prior diffusion-based baselines. The main limitations of this work are the limited dataset scale and its focus on unconditional generation. Future work should explore larger datasets and extend the method to conditional generation tasks.

\newpage
\newpage
\bibliographystyle{plain}
\bibliography{bibentries, references}

@misc{nie_large_2025,
	title = {Large {Language} {Diffusion} {Models}},
	url = {http://arxiv.org/abs/2502.09992},
	doi = {10.48550/arXiv.2502.09992},
	urldate = {2025-06-28},
	publisher = {arXiv},
	author = {Nie, Shen and Zhu, Fengqi and You, Zebin and Zhang, Xiaolu and Ou, Jingyang and Hu, Jun and Zhou, Jun and Lin, Yankai and Wen, Ji-Rong and Li, Chongxuan},
	  
	year = {2025},
	note = {arXiv:2502.09992 [cs]},
}

@misc{shaul_flow_2024,
	title = {Flow {Matching} with {General} {Discrete} {Paths}: {A} {Kinetic}-{Optimal} {Perspective}},
	shorttitle = {Flow {Matching} with {General} {Discrete} {Paths}},
	url = {http://arxiv.org/abs/2412.03487},
	doi = {10.48550/arXiv.2412.03487},
	urldate = {2025-06-28},
	publisher = {arXiv},
	author = {Shaul, Neta and Gat, Itai and Havasi, Marton and Severo, Daniel and Sriram, Anuroop and Holderrieth, Peter and Karrer, Brian and Lipman, Yaron and Chen, Ricky T. Q.},
	  
	year = {2024},
	note = {arXiv:2412.03487 [cs]},
}

@misc{kingma_adam_2017,
	title = {Adam: {A} {Method} for {Stochastic} {Optimization}},
	shorttitle = {Adam},
	url = {http://arxiv.org/abs/1412.6980},
	doi = {10.48550/arXiv.1412.6980},
	urldate = {2025-06-25},
	publisher = {arXiv},
	author = {Kingma, Diederik P. and Ba, Jimmy},
	  
	year = {2017},
	note = {arXiv:1412.6980 [cs]},
}

@misc{elfwing_sigmoid-weighted_2017,
	title = {Sigmoid-{Weighted} {Linear} {Units} for {Neural} {Network} {Function} {Approximation} in {Reinforcement} {Learning}},
	url = {http://arxiv.org/abs/1702.03118},
	doi = {10.48550/arXiv.1702.03118},
	urldate = {2025-06-25},
	publisher = {arXiv},
	author = {Elfwing, Stefan and Uchibe, Eiji and Doya, Kenji},
	  
	year = {2017},
	note = {arXiv:1702.03118 [cs]},
}

@article{maruyama_continuous_1955,
	title = {Continuous {Markov} processes and stochastic equations},
	volume = {4},
	issn = {1973-4409},
	url = {https://doi.org/10.1007/BF02846028},
	doi = {10.1007/BF02846028},
	language = {en},
	number = {1},
	urldate = {2025-06-22},
	journal = {Rendiconti del Circolo Matematico di Palermo},
	author = {Maruyama, Gisiro},
	  
	year = {1955},
	pages = {48--90},
}

@article{anderson_reverse-time_1982,
	title = {Reverse-time diffusion equation models},
	volume = {12},
	issn = {0304-4149},
	url = {https://www.sciencedirect.com/science/article/pii/0304414982900515},
	doi = {10.1016/0304-4149(82)90051-5},
	number = {3},
	urldate = {2025-06-03},
	journal = {Stochastic Processes and their Applications},
	author = {Anderson, Brian D. O.},
	  
	year = {1982},
	pages = {313--326},
}

@misc{perez_film_2017,
	title = {{FiLM}: {Visual} {Reasoning} with a {General} {Conditioning} {Layer}},
	shorttitle = {{FiLM}},
	url = {http://arxiv.org/abs/1709.07871},
	doi = {10.48550/arXiv.1709.07871},
	urldate = {2025-06-01},
	publisher = {arXiv},
	author = {Perez, Ethan and Strub, Florian and Vries, Harm de and Dumoulin, Vincent and Courville, Aaron},
	  
	year = {2017},
	note = {arXiv:1709.07871 [cs]},
}

@misc{sahoo_diffusion_2024,
	title = {Diffusion {Models} {With} {Learned} {Adaptive} {Noise}},
	url = {http://arxiv.org/abs/2312.13236},
	doi = {10.48550/arXiv.2312.13236},
	urldate = {2025-03-19},
	publisher = {arXiv},
	author = {Sahoo, Subham Sekhar and Gokaslan, Aaron and Sa, Chris De and Kuleshov, Volodymyr},
	  
	year = {2024},
	note = {arXiv:2312.13236 [cs]},
}

@article{lovelace2023latent,
  title={Latent diffusion for language generation},
  author={Lovelace, Justin and Kishore, Varsha and Wan, Chao and Shekhtman, Eliot and Weinberger, Kilian Q},
  journal={Advances in Neural Information Processing Systems},
  volume={36},
  pages={56998--57025},
  year={2023}
}

@inproceedings{pillutla_mauve_2021,
	title = {{MAUVE}: {Measuring} the {Gap} {Between} {Neural} {Text} and {Human} {Text} using {Divergence} {Frontiers}},
	volume = {34},
	shorttitle = {{MAUVE}},
	url = {https://proceedings.neurips.cc/paper/2021/hash/260c2432a0eecc28ce03c10dadc078a4-Abstract.html},
	urldate = {2025-03-07},
	booktitle = {Advances in {Neural} {Information} {Processing} {Systems}},
	publisher = {Curran Associates, Inc.},
	author = {Pillutla, Krishna and Swayamdipta, Swabha and Zellers, Rowan and Thickstun, John and Welleck, Sean and Choi, Yejin and Harchaoui, Zaid},
	year = {2021},
	pages = {4816--4828},
}

@misc{mostafazadeh_corpus_2016,
	title = {A {Corpus} and {Evaluation} {Framework} for {Deeper} {Understanding} of {Commonsense} {Stories}},
	url = {http://arxiv.org/abs/1604.01696},
	doi = {10.48550/arXiv.1604.01696},
	urldate = {2025-03-07},
	publisher = {arXiv},
	author = {Mostafazadeh, Nasrin and Chambers, Nathanael and He, Xiaodong and Parikh, Devi and Batra, Dhruv and Vanderwende, Lucy and Kohli, Pushmeet and Allen, James},
	  
	year = {2016},
	note = {arXiv:1604.01696 [cs]},
}

@misc{bartosh_neural_2024,
	title = {Neural {Diffusion} {Models}},
	url = {http://arxiv.org/abs/2310.08337},
	doi = {10.48550/arXiv.2310.08337},
	urldate = {2025-03-06},
	publisher = {arXiv},
	author = {Bartosh, Grigory and Vetrov, Dmitry and Naesseth, Christian A.},
	  
	year = {2024},
	note = {arXiv:2310.08337 [cs]},
}

@misc{devlin_bert_2019,
	title = {{BERT}: {Pre}-training of {Deep} {Bidirectional} {Transformers} for {Language} {Understanding}},
	shorttitle = {{BERT}},
	url = {http://arxiv.org/abs/1810.04805},
	doi = {10.48550/arXiv.1810.04805},
	urldate = {2025-03-06},
	publisher = {arXiv},
	author = {Devlin, Jacob and Chang, Ming-Wei and Lee, Kenton and Toutanova, Kristina},
	  
	year = {2019},
	note = {arXiv:1810.04805 [cs]},
}

@article{hoogeboom_autoregressive_2022,
  title={Autoregressive diffusion models},
  author={Hoogeboom, Emiel and Gritsenko, Alexey A and Bastings, Jasmijn and Poole, Ben and Berg, Rianne van den and Salimans, Tim},
  journal={arXiv preprint arXiv:2110.02037},
  year={2021}
}

@misc{shabalin_tencdm_2025,
	title = {{TEncDM}: {Understanding} the {Properties} of the {Diffusion} {Model} in the {Space} of {Language} {Model} {Encodings}},
	shorttitle = {{TEncDM}},
	url = {http://arxiv.org/abs/2402.19097},
	doi = {10.48550/arXiv.2402.19097},
	urldate = {2025-03-04},
	publisher = {arXiv},
	author = {Shabalin, Alexander and Meshchaninov, Viacheslav and Chimbulatov, Egor and Lapikov, Vladislav and Kim, Roman and Bartosh, Grigory and Molchanov, Dmitry and Markov, Sergey and Vetrov, Dmitry},
	  
	year = {2025},
	note = {arXiv:2402.19097 [cs]},
}

@misc{li_diffusion-lm_2022,
	title = {Diffusion-{LM} {Improves} {Controllable} {Text} {Generation}},
	url = {http://arxiv.org/abs/2205.14217},
	doi = {10.48550/arXiv.2205.14217},
	urldate = {2025-01-08},
	publisher = {arXiv},
	author = {Li, Xiang Lisa and Thickstun, John and Gulrajani, Ishaan and Liang, Percy and Hashimoto, Tatsunori B.},
	  
	year = {2022},
	note = {arXiv:2205.14217 [cs]},
}

@misc{dieleman_continuous_2022,
	title = {Continuous diffusion for categorical data},
	url = {http://arxiv.org/abs/2211.15089},
	doi = {10.48550/arXiv.2211.15089},
	urldate = {2025-01-07},
	publisher = {arXiv},
	author = {Dieleman, Sander and Sartran, Laurent and Roshannai, Arman and Savinov, Nikolay and Ganin, Yaroslav and Richemond, Pierre H. and Doucet, Arnaud and Strudel, Robin and Dyer, Chris and Durkan, Conor and Hawthorne, Curtis and Leblond, Rémi and Grathwohl, Will and Adler, Jonas},
	  
	year = {2022},
	note = {arXiv:2211.15089 [cs]},
}

@misc{gao_empowering_2024,
	title = {Empowering {Diffusion} {Models} on the {Embedding} {Space} for {Text} {Generation}},
	url = {http://arxiv.org/abs/2212.09412},
	doi = {10.48550/arXiv.2212.09412},
	urldate = {2025-01-07},
	publisher = {arXiv},
	author = {Gao, Zhujin and Guo, Junliang and Tan, Xu and Zhu, Yongxin and Zhang, Fang and Bian, Jiang and Xu, Linli},
	  
	year = {2024},
	note = {arXiv:2212.09412 [cs]},
}

@misc{peebles_scalable_2023,
	title = {Scalable {Diffusion} {Models} with {Transformers}},
	url = {http://arxiv.org/abs/2212.09748},
	doi = {10.48550/arXiv.2212.09748},
	urldate = {2025-01-06},
	publisher = {arXiv},
	author = {Peebles, William and Xie, Saining},
	  
	year = {2023},
	note = {arXiv:2212.09748 [cs]},
}

@misc{campbell_continuous_2022,
	title = {A {Continuous} {Time} {Framework} for {Discrete} {Denoising} {Models}},
	url = {http://arxiv.org/abs/2205.14987},
	doi = {10.48550/arXiv.2205.14987},
	urldate = {2025-01-03},
	author = {Campbell, Andrew and Benton, Joe and Bortoli, Valentin De and Rainforth, Tom and Deligiannidis, George and Doucet, Arnaud},
	  
	year = {2022},
	note = {arXiv:2205.14987 [stat]},
}

@misc{song_score-based_2021,
	title = {Score-{Based} {Generative} {Modeling} through {Stochastic} {Differential} {Equations}},
	url = {http://arxiv.org/abs/2011.13456},
	urldate = {2024-11-13},
	publisher = {arXiv},
	author = {Song, Yang and Sohl-Dickstein, Jascha and Kingma, Diederik P. and Kumar, Abhishek and Ermon, Stefano and Poole, Ben},
	  
	year = {2021},
	note = {arXiv:2011.13456},
}

@inproceedings{ho_denoising_2020,
	title = {Denoising {Diffusion} {Probabilistic} {Models}},
	volume = {33},
	url = {https://proceedings.neurips.cc/paper_files/paper/2020/hash/4c5bcfec8584af0d967f1ab10179ca4b-Abstract.html},
	urldate = {2024-11-06},
	booktitle = {Advances in {Neural} {Information} {Processing} {Systems}},
	publisher = {Curran Associates, Inc.},
	author = {Ho, Jonathan and Jain, Ajay and Abbeel, Pieter},
	year = {2020},
	pages = {6840--6851},
}

@misc{kingma_variational_2023,
	title = {Variational {Diffusion} {Models}},
	url = {http://arxiv.org/abs/2107.00630},
	urldate = {2024-11-02},
	publisher = {arXiv},
	author = {Kingma, Diederik P. and Salimans, Tim and Poole, Ben and Ho, Jonathan},
	  
	year = {2023},
	note = {arXiv:2107.00630},
}

@misc{song_denoising_2022,
	title = {Denoising {Diffusion} {Implicit} {Models}},
	url = {http://arxiv.org/abs/2010.02502},
	urldate = {2024-10-31},
	publisher = {arXiv},
	author = {Song, Jiaming and Meng, Chenlin and Ermon, Stefano},
	  
	year = {2022},
	note = {arXiv:2010.02502},
}

@misc{gat_discrete_2024,
	title = {Discrete {Flow} {Matching}},
	url = {http://arxiv.org/abs/2407.15595},
	doi = {10.48550/arXiv.2407.15595},
	urldate = {2024-10-28},
	publisher = {arXiv},
	author = {Gat, Itai and Remez, Tal and Shaul, Neta and Kreuk, Felix and Chen, Ricky T. Q. and Synnaeve, Gabriel and Adi, Yossi and Lipman, Yaron},
	year = {2024},
	note = {arXiv:2407.15595},
}

@article{gulrajani_likelihood-based_nodate,
  title={Likelihood-based diffusion language models},
  author={Gulrajani, Ishaan and Hashimoto, Tatsunori B},
  journal={Advances in Neural Information Processing Systems},
  volume={36},
  pages={16693--16715},
  year={2023}
}

@inproceedings{bartosh_neural_flow_2024,
 author = {Bartosh, Grigory and Vetrov, Dmitry and Naesseth, Christian A.},
 booktitle = {Advances in Neural Information Processing Systems},
 editor = {A. Globerson and L. Mackey and D. Belgrave and A. Fan and U. Paquet and J. Tomczak and C. Zhang},
 pages = {73952--73985},
 publisher = {Curran Associates, Inc.},
 title = {Neural Flow Diffusion Models: Learnable Forward Process for Improved Diffusion Modelling},
 url = {https://proceedings.neurips.cc/paper_files/paper/2024/file/871a8ccb9232487366feb5e2d9069915-Paper-Conference.pdf},
 volume = {37},
 year = {2024}
}

@misc{jordan2024muon,
  author       = {Keller Jordan and Yuchen Jin and Vlado Boza and You Jiacheng and
                  Franz Cesista and Laker Newhouse and Jeremy Bernstein},
  title        = {Muon: An optimizer for hidden layers in neural networks},
  year         = {2024},
  url          = {https://kellerjordan.github.io/posts/muon/}
}

@article{radford2019language,
  title={Language models are unsupervised multitask learners},
  author={Radford, Alec and Wu, Jeffrey and Child, Rewon and Luan, David and Amodei, Dario and Sutskever, Ilya and others},
  journal={OpenAI blog},
  volume={1},
  number={8},
  pages={9},
  year={2019}
}

@misc{google_deepmind_2025,
  author       = {Google DeepMind},
  title        = {Gemini Diffusion},
  year         = {2025},
  howpublished = {\url{https://blog.google/technology/google-deepmind/gemini-diffusion/}},
  note         = {Accessed: 2025-10-11}
}

@inproceedings{chen_cheaper_2023,
    address = {Singapore},
    title = {A {Cheaper} and {Better} {Diffusion} {Language} {Model} with {Soft}-{Masked} {Noise}},
    url = {https://aclanthology.org/2023.emnlp-main.289/},
    doi = {10.18653/v1/2023.emnlp-main.289},
    urldate = {2025-06-28},
    booktitle = {Proceedings of the 2023 {Conference} on {Empirical} {Methods} in {Natural} {Language} {Processing}},
    publisher = {Association for Computational Linguistics},
    author = {Chen, Jiaao and Zhang, Aston and Li, Mu and Smola, Alex and Yang, Diyi},
    editor = {Bouamor, Houda and Pino, Juan and Bali, Kalika},

    year = {2023},
    pages = {4765--4775},
}


\appendix

\section{Details of NFDM} \label{app: nfdm exact}
NFDM \cite{bartosh_neural_flow_2024} extends the flexibility of the forward process by allowing its parameters to be conditioned directly on the input $\bm{x}$. In this formulation, the mean and variance of the forward process are parameterized via nonlinear functions of $\bm{x}$ and $t$:

\begin{align} \label{NFDM-F}
F^{\text{NFDM}}_{\varphi}(\epsilon, t, E_\varphi({\bm{x}})) &= \mu_{\varphi}(E_\varphi({\bm{x}}), t) + \sigma_{\varphi}(E_\varphi({\bm{x}}), t) \epsilon. \\
 \mu_{\varphi}(E_\varphi({\bm{x}}), t) &= (1 - t)\, E_\varphi({\bm{x}}) \;+\; t(1 - t)\,\overline{\mu}_{\varphi}( E_\varphi({\bm{x}}), t), \\[1ex]
\sigma_{\varphi}(E_\varphi({\bm{x}}), t) &= \delta^{\,1 - t}\,\bigl(\overline{\sigma}_{\varphi}(E_\varphi({\bm{x}}), t))^{\,t(1 - t)}.
\end{align}

At the boundaries, the mean and variance functions satisfy
\begin{align}
    \mu_{\varphi}(E_{\varphi}(\bm{x}), 0) &= E_{\varphi}(\bm{x}), &
    \sigma_{\varphi}(E_{\varphi}(\bm{x}), 0) &= \delta, \\
    \mu_{\varphi}(E_{\varphi}(\bm{x}), 1) &= 0, &
    \sigma_{\varphi}(E_{\varphi}(\bm{x}), 1) &= 1,
\end{align}
where we set $\delta = 0.01$. These boundary conditions ensure that
\begin{align}
    q_{\varphi}(\bm{z}_0 \mid \bm{x}) &= 
    \mathcal{N}\big(\bm{z}_0; E_{\varphi}(\bm{x}), \delta^2 I\big), &
    q_{\varphi}(\bm{z}_1 \mid \bm{x}) &= 
    \mathcal{N}\big(\bm{z}_1; 0, I\big).
\end{align}
For intermediate $0 < t < 1$, $\mu_{\varphi}$ and $\sigma_{\varphi}$ can take arbitrary values, defining a family of conditional Gaussian marginals

NFDM learns a model that can generate samples from the data distribution by parameterizing the reverse dynamics with a neural network that predicts the embeddings $\hat{E}_\theta(\bm{z}_t, t)$. We can learn the parameters of this predictor $\theta$, jointly with the parameters of the forward process $\varphi$. The loss used to achieve this is derived in \citep{bartosh_neural_flow_2024} and shown in \autoref{NFDM elbo latent} for completeness. 


As explained in the Background \autoref{sec:related_work} NFDM gets a generative model by learning to reverse the forward conditional SDE. This is done by first introducing the reverse conditional SDE, and then learning a generative reverse process parameterized by a prediction of the embeddings. The drift terms of the forward and reverse conditional SDEs both include $f(z_t,t, E_{\varphi}(\bm{x}))$. This is the ODE drift obtained by differentiating the trajectories from $\bm{z}_0$ to $\bm{z}_1$ induced by $F_\varphi$, over time. This leads to a velocity field corresponding to the conditional distribution, thereby defining a conditional ODE:
\begin{equation}
    d\bm{z}_t = f(z_t,t, E_{\varphi}(\bm{x}))dt, \quad \text{where} \quad f(z_t,t, E_{\varphi}(\bm{x})) = \frac{\partial F_{\varphi}(\epsilon, t, E_{\varphi}(\bm{x}))}{\partial t} \bigg |_{\epsilon =  F_{\varphi}^{-1}(\bm{z}_t, t, E_{\varphi}(\bm{x}))}
\end{equation}

\noindent As explained in \autoref{sec:related_work}, the (conditional) distribution of the latent variable trajectories can then be described by an initial distribution $q(z_0\!\mid E_{\varphi}(\bm{x}))$ and a Stochastic Differential Equation (SDE) with a drift term $\tilde{f}^{F}:\mathbb{R}^{D}\!\times[0,1]\!\to\!\mathbb{R}^{D}$, scalar variance $g(t):[0,1]\!\to\!\mathbb{R}_{+}$, and a standard Wiener process $\bm{w}$:
\begin{align} \label{forwardsde}
    \mathrm{d}\bm{z}_t &= \tilde{f}^{F}(\bm{z}_t,t, E_{\varphi}(\bm{x}))\,dt + g^{2}_{\varphi}(t)\,d\bm{w}, \quad \text{where} \\
    \tilde{f}^{F}(\bm{z}_t,t,\bm{x})&=f(\bm{z}_t,t, E_{\varphi}(\bm{x}))+\frac{g^{2}_{\varphi}(t)}{2}\,\nabla_{\bm{z}_t}\log q_{\varphi}(\bm{z}_t|E_{\varphi}(\bm{x})).
\end{align}

\noindent To create a target for our learned generative process, this forward process is then reversed by starting from the prior $z_1\sim\mathcal{N}(z_1;0,I)$ and following the reverse SDE \cite{anderson_reverse-time_1982}:
\begin{align} \label{reversesde}
    \mathrm{d}\bm{z}_t &= \tilde{f}^{B}(\bm{z}_t,t,E_{\varphi}(\bm{x}))\,dt + g(t)\,d\bar{\bm{w}},  \quad \text{where} \\
    \tilde{f}^{B}(\bm{z}_t,t,\bm{x})&=f(\bm{z}_t,t, E_{\varphi}(\bm{x}))-\frac{g^{2}_{\varphi}(t)}{2}\,\nabla_{\bm{z}_t}\log q_{\varphi}(z_t|E_{\varphi}(\bm{x})).
\end{align}
Here, $\bar{\bm{w}}$ denotes a standard Wiener process where time flows backward. This reversal results in a sample $\bm{z}_0 \sim p_\theta(\bm{z_0})$, such that decoding $\bm{z}_0$ through dot product similarity search with the vocabulary embeddings leads to an approximate sample from the data distribution $q(\bm{x})$. The reverse and forward processes can be learned jointly through the ELBO,
 
\begin{align} \label{NFDM elbo latent}
&\mathbb{E}_{q(\bm{x})} [-\log \tilde{p}_{\theta, \varphi}(\bm{x})] 
\leq \tilde{\mathcal{L}}_{\text{rec}} + \tilde{\mathcal{L}}_{\text{diff}} + \tilde{\mathcal{L}}_{\text{prior}}, \qquad \text{where} \\
&\tilde{\mathcal{L}}_{\text{rec}} = \mathbb{E}_{q_{\varphi}(\bm{x}, \bm{z}_0)} 
[-\log p(\bm{x} | \bm{z}_0)], \\
&\tilde{\mathcal{L}}_{\text{diff}} = \mathbb{E}_{u(t) \, q(\bm{x}) \, q_{\varphi}(\bm{z}_t \mid \bm{x})} 
\Bigg[
    \frac{1}{2 g_{\varphi}^2(t)}
    \big\|
        \tilde{f}_{\varphi}^{B}(\bm{z}_t, t, \bm{x})) 
        - \hat{f}_{\theta, \varphi}^{B}(\bm{z}_t, t)
    \big\|_2^2
\Bigg]\\
&\tilde{\mathcal{L}}_{\text{prior}} = \mathbb{E}_{q(\bm{x})} 
[\mathcal{D}_{\text{KL}} (q_{\varphi}(\bm{z}_1 | \bm{x}) 
\parallel \tilde{p}(\bm{z}_1))].
\end{align}

Here, the reverse drift is parameterized as,

\begin{equation} \label{reverse_flow_hat}
     \hat{f}_{\theta, \varphi}^{B}(\bm{z}_t, t) = f(\bm{z}_t,t, \hat{E}_\theta(\bm{z}_t, t))-\frac{g^{2}_{\varphi}(t)}{2}\,\nabla_{\bm{z}_t}\log q_{\varphi}(\bm{z}_t|\hat{E}_\theta(\bm{z}_t, t))
\end{equation}

\section{Details of MuLAN} \label{details of mulan}

In this section, we derive a simplified version of the NFDM loss that holds when the forward process is parameterized as in MuLAN \cite{sahoo_diffusion_2024}. We derive this loss for the version of MuLAN that uses the auxiliary latent variable $\bm{c}$. The same derivations also when we omit the conditioning. Note that in the derivations below, multiplication and exponentiation of $\bm{\alpha}$, $\bm{\sigma}$ and $\bm\gamma$ are always elementwise. 

\subsection{MuLAN forward process}

The forward process for MuLAN is defined through $F^{MuLAN}_\varphi(\epsilon, t,  E_\varphi({\bm{x}}), \bm{c})$:

\begin{equation}
F^{MuLAN}_\varphi(\epsilon, t,  E_\varphi({\bm{x}}), \bm{c}) = \bm{\alpha}_{\varphi}(t,\bm{c}) E_\varphi({\bm{x}}) + \bm{\sigma}_{\varphi}(t,\bm{c}) \epsilon
\end{equation}

\noindent Since in MuLAN we learn the SNR, this means that we have:

\begin{align}
\bm{z}_t = \bm{\alpha}_{\varphi}(t,\bm{c}) E_\varphi({\bm{x}}) + \bm{\alpha}_{\varphi}(t,\bm{c}) \epsilon 
\quad \text{where} \quad
\bm{\alpha}_{\varphi}(t,\bm{c})^2 + \bm{\sigma}_{\varphi}(t,\bm{c})^2 = 1
\end{align}

\noindent For simplicity, we define $\bm{\alpha} \equiv \bm{\alpha}_{\varphi}(t,\bm{c})$ and $\bm{\sigma} = \bm{\sigma}_{\varphi}(t,\bm{c})$ from here onwards. Additionally, we define $\bm{x} \equiv E_\varphi({\bm{x}})$, since these derivations hold both in embedding space and when directly working with a continuous modality. Note that we can not have $\bm{x}$ reflect a discrete sequence like a sequence of tokens. In that case, we have to use the embeddings. Some simple rewriting can then give us the following connections:

\begin{align}
    \bm{x} = \frac{\bm{z}_t - \bm{\alpha} \varepsilon}{\bm{\alpha}}
    \quad \text{and} \quad
    \varepsilon = \frac{\bm{z}_t - \bm{\alpha} \bm{x}}{\bm{\sigma}}
\end{align}

As mentioned above, MuLAN learns the forward process through the Signal-To-Noise Ratio (SNR):

\begin{equation} \label{Mulan SNR}
    \textbf{SNR}(t) = \bm{\alpha}^2/\bm{\sigma}^2 = e^{-\bm{\gamma}_\varphi(t, \bm{c})}
\end{equation}

\noindent For simplicity we use $\bm{\gamma} \equiv \bm{\gamma}_\varphi(t,\bm{c})$ from this point forward. $\bm{\gamma}$ is the quantity that we learn using a monotonically increasing network. Then we can rewrite the $\bm{\alpha}$ and $\bm{\sigma}$ coefficients in terms of the gamma function,

\begin{align} \label{alpha gamma connection}
    SNR = \frac{\bm\alpha^2}{1 - \bm\alpha^2} = e^{-\bm\gamma}
    \quad \Rightarrow \quad
    \bm\alpha^2 &= \frac{e^{-\bm\gamma}}{1 + e^{-\bm\gamma}} = \frac{1}{1 + e^{\bm\gamma}} = \text{sigmoid}(-\bm\gamma) \\
    \bm\sigma^2 &= 1 - \bm\alpha^2 = \frac{1}{1 + e^{-\bm\gamma}} = \text{sigmoid}(\bm\gamma)
\end{align}

The NFDM loss is defined in terms of the reverse SDE. This reverse SDE consists of the conditional ODE, the score function and volatility with Brownian motion. The conditional ODE is:

\begin{align}
    f = \dot{\bm\alpha} x + \dot{\bm\sigma} \varepsilon = \dot{\bm\alpha} x + \frac{\dot{\bm\sigma}}{\bm\sigma} (\bm{z}_t - \bm\alpha \bm{x})
\end{align}

\noindent The conditional score function is:

\begin{align}
    s = - \frac{\varepsilon}{\bm\sigma} = \frac{\bm\alpha \bm{x} - \bm{z}_t}{\bm\sigma^2}
\end{align}

\noindent Combining the drift of the ODE $f$ and the score function $s$ with the volatility $\bm{g}(t)$ we can write down the conditional forward SDE:

\begin{align}
    d z = f^F d t + \bm{g}(t) d w, \quad \text{where} \quad f^F = f + \frac{\bm{g}^2(t)}{2} s
\end{align}

\noindent Similarly, we can write down the conditional backwards SDE:

\begin{align}
    d z = f^B d t + \bm{g}(t) d \bar{w}, \quad \text{where} \quad f^B = f - \frac{\bm{g}^2(t)}{2} s
\end{align}

\subsection{Markovian volatility}

In general, volatility $\bm{g}(t)$ can be an arbitrary (positive) function of time $t$. However, there is one useful consideration that can help us parameterise more efficiently.

In diffusion models, our aim is to match the distribution of trajectories of the forward and reversed processes. The reverse process is Markovian by design. Therefore, to be able to match the distributions of trajectories, the forward process should also be Markovian. To guarantee this, we can find such a volatility $\bm{g}(t)$ that makes the forward process independent of $\bm{x}$. In the case of MuLAN and also VDM, this can be done analytically. 

\begin{align}
    f^F
    &= f + \frac{\bm{g}(t)^2}{2} s \\
    &= \dot{\bm\alpha} x + \frac{\dot{\bm\sigma}}{\bm\sigma} (\bm{z}_t - \bm\alpha \bm{x}) + \frac{\bm{g}^2(t)}{2} \frac{\bm\alpha \bm{x} - \bm{z}_t}{\bm\sigma^2} \\
    &= \underbrace{ \left( \dot{\bm\alpha} - \frac{\dot{\bm\sigma}}{\bm\sigma} \bm\alpha + \frac{\bm{g}^2(t)}{2} \frac{\bm\alpha}{\bm\sigma^2} \right) }_{=0} \bm{x} + \left( \frac{\dot{\bm\sigma}}{\bm\sigma} - \frac{\bm{g}^2(t)}{2} \frac{1}{\bm\sigma^2} \right) \bm{z}_t \\
\end{align}

\noindent Setting the final term on the left equal to zero gives us an expression for the volatility $\bm{g}(t)$.

\begin{align}
    \bm{g}^2(t)
    &= 2 \frac{\bm\sigma^2}{\bm\alpha} \left( \frac{\dot{\bm\sigma}}{\bm\sigma} \bm\alpha - \dot{\bm\alpha} \right) \\
    &= 2 \bm\sigma \dot{\bm\sigma} - 2 \bm\sigma^2 \frac{\dot{\bm\alpha}}{\bm\alpha} \\
    &= (\bm\sigma^2)' - 2 (\log \bm\alpha)' \bm\sigma^2
\end{align}

\noindent We can also rewrite the volatility in terms of the gamma function. 

\begin{align}
    \bm{g}^2(t)
    &= (\bm\sigma^2)' - 2 (\log \bm\alpha)' \bm\sigma^2 \\
    &= (\bm\sigma^2)' - \frac{2 \bm\alpha \dot{\bm\alpha}}{\bm\alpha^2} \bm\sigma^2 \\
    &= (\bm\sigma(\bm\gamma))' - \frac{(\bm\sigma(-\bm\gamma))'}{\bm\sigma(-\bm\gamma)} \bm\sigma(\bm\gamma) \\
    &= \bm\sigma(\bm\gamma) \left( 1 - \bm\sigma(\bm\gamma) \right) \dot{\bm\gamma} + \frac{\bm\sigma(-\bm\gamma) \left( 1 - \bm\sigma(-\bm\gamma) \right) \dot{\bm\gamma}}{\bm\sigma(-\bm\gamma)} \bm\sigma(\bm\gamma) \\
    &= \bm\sigma(\bm\gamma) \dot{\bm\gamma} \left( 1 - \bm\sigma(\bm\gamma) + \underbrace{1 - \bm\sigma(-\bm\gamma)}_{=\bm\sigma(\bm\gamma)} \right) \\
    &= \bm\sigma(\bm\gamma) \dot{\bm\gamma}
\end{align}

\noindent To keep the volatility function general, but preserve the connection with the gamma function, we can reparameterize the volatility function as follows:

\begin{align}
    \bm{g}^2(t) = \bm\sigma(\bm\gamma) \dot{\bm\gamma} \eta
\end{align}

\noindent where $\eta$ is an arbitrary non-negative function of time $t$. If we set $\eta = 1$, we will recover the Markovian volatility.

\subsection{Simplified MuLAN ELBO}

When using the Markovian parameterisation of the volatility and the Gaussian forward process used in Mulan, we can significantly simplify the ELBO. Doing this allows us to reduce the numerical error that can occur when multiplying many small values, and provides us with a more interpretable objective.
We define the reverse process through prediction $\hat{\bm{x}}(\bm{z}_t,t)$ that we substitute into the conditional backward SDE:

\begin{align}
    d z = \hat{f}^B d t + \bm{g}(t) d \bar{w}, \quad \text{where} \quad \hat{f}^B(z, t) = f^B(z, t, \hat{\bm{x}}(\bm{z}_t,t))
\end{align}

\noindent We know that the diffusion loss in NFDM is defined as:

\begin{align}
    \mathcal{L}_{\text{diff}} = \mathbb{E}_{u(t) \, q(\bm{x}) \, q_{\varphi}(\bm{z}_t \mid \bm{x})}\bigg[ \lambda_{f^B}  \left\| f^B - \hat{f}^B \right\|_2^2 \bigg], \quad \text{where} \quad \lambda_{f^B} = \frac{1}{2 \bm{g}^2(t)}
\end{align}

\noindent For MuLAN, we can rewrite the $f^B$ as:
\begin{align}
    f^B
    &= f - \frac{\bm{g}^2(t)}{2} s \\
    &= \dot{\bm\alpha} x + \frac{\dot{\bm\sigma}}{\bm\sigma} (\bm{z}_t - \bm\alpha \bm{x}) - \frac{\bm{g}^2(t)}{2} \frac{\bm\alpha \bm{x} - \bm{z}_t}{\bm\sigma^2} \\
    &= \left( \dot{\bm\alpha} - \frac{\dot{\bm\sigma}}{\bm\sigma} \bm\alpha - \frac{\bm{g}^2(t)}{2} \frac{\bm\alpha}{\bm\sigma^2} \right) x + \left( \frac{\dot{\bm\sigma}}{\bm\sigma} + \frac{\bm{g}^2(t)}{2} \frac{1}{\bm\sigma^2} \right) \bm{z}_t
\end{align}

\noindent Since the second term doesn't depend on $\bm{x}$ and will cancel out in the ELBO, we can rewrite the ELBO as:

\begin{align}
    \mathcal{L} = \mathbb{E}_{u(t) \, q(\bm{x}) \, q_{\varphi}(\bm{z}_t \mid \bm{x})}\bigg[ \lambda_x \left\| \bm{x} - \hat{\bm{x}} \right\|_2^2 \bigg]
\end{align}

\noindent We can now derive the $\lambda_x$ coefficient,

\begin{align}
    \lambda_x
    &= \frac{1}{2 \bm{g}^2(t)} \left( \dot{\bm\alpha} - \frac{\dot{\bm\sigma}}{\bm\sigma} \bm\alpha - \frac{\bm{g}^2(t)}{2} \frac{\bm\alpha}{\bm\sigma^2} \right)^2 \\
    &= \frac{1}{2 \bm{g}^2(t)} \left( \frac{\bm\alpha}{2} \frac{2 \bm\alpha \dot{\bm\alpha}}{\bm\alpha^2} - \frac{\bm\alpha}{2} \frac{2 \bm\sigma \dot{\bm\sigma}}{\bm\sigma^2} - \frac{\bm\alpha}{2} \frac{g^2}{\bm\sigma^2} \right)^2 \\
    &= \frac{1}{2 \bm{g}^2(t)} \frac{\bm\alpha^2}{2^2} \left( \frac{(\bm\alpha^2)'}{\bm\alpha^2} - \frac{(\bm\sigma^2)'}{\bm\sigma^2} - \frac{g^2}{\bm\sigma^2} \right)^2 \\
    &= \frac{1}{2} \frac{\bm\alpha^2}{\bm\sigma^2} \frac{1}{2^2 \dot{\bm\gamma} \eta} \left( \frac{(\bm\alpha^2)'}{\bm\alpha^2} - \frac{(\bm\sigma^2)'}{\bm\sigma^2} - \frac{\bm\sigma^2 \dot{\bm\gamma} \eta}{\bm\sigma^2} \right)^2 \\
    &= \frac{1}{2} e^{-\bm\gamma} \frac{1}{2^2 \dot{\bm\gamma} \eta} \left( \frac{\bm\sigma(-\bm\gamma) \left( 1 - \bm\sigma(-\bm\gamma) \right) (-1) \dot{\bm\gamma}}{\bm\sigma(-\bm\gamma)} - \frac{\bm\sigma(\bm\gamma) \left( 1 - \bm\sigma(\bm\gamma) \right) \dot{\bm\gamma}}{\bm\sigma(\bm\gamma)} - \dot{\bm\gamma} \eta \right)^2 \\
    &= \frac{1}{2} e^{-\bm\gamma} \frac{1}{2^2 \dot{\bm\gamma} \eta} \left( \big[ - \underbrace{\left( 1 - \bm\sigma(-\bm\gamma) \right)}_{=\bm\sigma(\bm\gamma)} -  1 + \bm\sigma(\bm\gamma) \big] \dot{\bm\gamma} - \dot{\bm\gamma} \eta \right)^2 \\
    &= \frac{1}{2} e^{-\bm\gamma} \frac{1}{2^2 \dot{\bm\gamma} \eta} \left( - \dot{\bm\gamma} - \dot{\bm\gamma} \eta \right)^2 \\
    &= \frac{1}{2} e^{-\bm\gamma} \frac{ \dot{\bm\gamma}^2 \left( 1 + \eta \right)^2 }{2^2 \dot{\bm\gamma} \eta} \\
    &= \frac{1}{2} e^{-\bm\gamma} \dot{\bm\gamma} \frac{ \left( 1 + \eta \right)^2 }{2^2 \eta}
\end{align}

\noindent Since nothing except the last coefficient depends on function $\eta$. By computing the first and second derivatives of this term we can find the optimal $\eta$. It is $\eta = 1$. Therefore, the optimal volatility function is a Markovian volatility. We can also find a connection with the SNR function, when $\eta = 1$.

\begin{align}
    \text{SNR}' = (e^{-\gamma})' = -e^{-\gamma} \dot{\gamma}, \quad \lambda_x = \frac{1}{2} e^{-\gamma} \dot{\gamma} = - \frac{1}{2} \text{SNR}'
\end{align}

\subsection{Fixed average SNR} \label{global SNR}
In \autoref{sec:method subsec:nfdminlatentspace}, we discussed restricting the flexibility of the learned forward process. Specifically, we use a restricted formulation of $\bm{\gamma}_\varphi(t, \bm{c})$ to prevent collapse when training with the rescaled ELBO.
\begin{align}
\bm{\gamma}_\varphi(t,\bm{c}))_j &= \gamma(t)^{\text{global}} + \bm{\gamma}'_\varphi(t,\bm{c})_j - \tilde{\gamma}_\varphi(t,c), \quad \text{where,} \\
\tilde{\gamma}_\varphi(t,c)' &= \log D - \log \sum_i^D \exp(-\bm{\gamma}'_\varphi(t,\bm{c})_i)
\end{align}

\noindent This parameterization fixes the global average of the SNR, over the spatial dimensions of gamma, to the SNR given by $\gamma(t)^{\text{global}}$. In our case, $\gamma(t)^{\text{global}}$ is fixed to the SNR of the baseline model Diffusion-LM. this SNR is derived in \autoref{App: diff-LM}. In the following, we show that this is the case by computing the average SNR over the dimensions D of $\bm{\gamma}_\varphi(\bm{c},t)$. Where D corresponds to the total dimensionality of $\bm{\gamma}_\varphi(\bm{c},t)$, multiplying the spatial dimensions. 

\begin{align}
    \frac{1}{D} \sum_{j=1}^{D} SNR(t)_j &= \frac{1}{D} \sum_{j=1}^{D} e^{-\gamma_\varphi(t, \bm{c})_j} \\
    &= \frac{1}{D} \sum_{j=1}^{D} e^{-\gamma(t)^{\text{global}} - \gamma'_\varphi(t,\bm{c})_j + \tilde{\gamma}_\varphi(t,\bm{c})} \\
    &= \frac{1}{D} \sum_{j=1}^{D} e^{-\gamma(t)^{\text{global}} - \gamma'_\varphi(t,\bm{c})_j + \log D - \log \sum_i e^{(-\gamma'_\varphi(t,\bm{c})_i)}} \\
    &= \frac{1}{D} \sum_{j=1}^{D} e^{-\gamma(t)^{\text{global}}}  e^{-\gamma'_\varphi(t,\bm{c})_j} D \frac{1}{\sum_i e^{(-\gamma'_\varphi(t,\bm{c})_i)}} \\
    &= e^{-\gamma(t)^{\text{global}}} \left(\sum_{j=1}^{D}   e^{-\gamma'_\varphi(t,\bm{c})_j}\right) \frac{1}{\sum_i^D e^{(-\gamma'_\varphi(t,\bm{c})_i)}} \\
    &= e^{-\gamma(t)^{\text{global}}} \\
\end{align}

\section{Continuous time Diffusion-LM} \label{App: diff-LM}
In this work, we use Diffusion-LM \cite{li_diffusion-lm_2022} as the baseline model with a static, non-learnable, forward process. We focus on using the continuous-time NFDM framework in this work. Since Diffusion-LM as presented in \cite{li_diffusion-lm_2022} we rewrite it to continuous time form using the NFDM framework in the section below. Note that as T approaches infinity, the discrete-time model matches the continuous-time model precisely. Since \cite{li_diffusion-lm_2022} uses $T=2000$ that means we saw very similar training losses in practice between the two models. 

The Diffusion-LM forward process is described by $q(\bm{z}_t \mid \bm{z}_{t-1}) = \mathcal{N}(\bm{z}_t; \sqrt{1 - \beta_t} \bm{z}_{t-1}, \beta_t \bm{I})$, where $\bm{z}_t$ are the continuous latent variables. The forward variance schedule is governed by $\bar{\alpha}_t = \prod_{s=1}^{t} (1 - \beta_s)$, with $\bar{\alpha}_t = 1 - \sqrt{t/T + s}$. This corresponds to the following parameterization of $F$ in the NFDM framework, 

\begin{align}
    F(x, t, \varepsilon) &= \alpha(t) E_\varphi({\bm{x}})  + \sigma(t) \varepsilon \quad \text{where,}\\
    \alpha(t) &= \sqrt{1 - \sqrt{t + s}} \\
    \sigma(t) &= \sqrt{\sqrt{t + s}}
\end{align}

\noindent In diffusion-LM, $s$ is a small constant set to $0.0001$. To ensure differentiability with respect to $t$, we set it to the following in NFDM,

\begin{equation} \label{s parameter}
    s = (0.99 - t) \cdot 0.0001
\end{equation}

\noindent We can also rewrite this in terms of the $\gamma(t)$ parameterization used in VDM \cite{kingma_variational_2023} and MuLAN \cite{sahoo_diffusion_2024}. This is necessary since we want to use the Diffusion-LM forward process as the reference process when training MuLAN with a rescaled loss. Using the connection of $\gamma(t)$ and $\alpha(t)$ in \autoref{alpha gamma connection} we get,

\begin{align}
    \alpha^2 &= \text{sigmoid}(-\gamma(t)) =  1 - \sqrt{t + s}\\
    &\Leftrightarrow \\
    \frac{1}{1+e^{\gamma(t)}} &= 1 - \sqrt{t + s} \\
    &\Leftrightarrow \\
     e^{\gamma(t)} &= \frac{1}{1 - \sqrt{t + s}} - 1 \\
    &\Leftrightarrow \\
    \gamma(t) &= \ln\left(\frac{1}{1 - \sqrt{t + s}} - 1\right) \\
    &= \ln\left(\frac{\sqrt{t + s}}{1 - \sqrt{t + s}}\right) \\
    &= \ln\left(\sqrt{t + s}\right) - \ln\left(1 - \sqrt{t + s}\right)
\end{align}

Note that we can arrive at the same conclusion by starting from the connection between $\gamma(t)$ and $\sigma(t)$ in this case. Additionally, in the used version of gamma, we clamp $\sqrt{t + s}$ between $1e^{-6}$ and $1-1e^{-6}$ for numerical stability in the fraction and natural logarithm. 

\noindent To match the continuous-time formulation of diffusion in NFDM to the discrete-time formulation in Diffusion-LM, we need to set the volatility $g^2(t)$ according to the following formula presented in \cite{kingma_variational_2023}, which allows us to convert from discrete to continuous time diffusion models.

\begin{align}
    g^2(t) &= \frac{d\sigma(t)^2}{dt} - 2 \frac{d \log \alpha(t)}{dt} \sigma_t^2 \\
    &= \frac{0.9999}{(2\sigma^2 (1-\sigma^2))}
\end{align}

\begin{figure}[H]
    \centering
    \includegraphics[width = 0.45\linewidth]{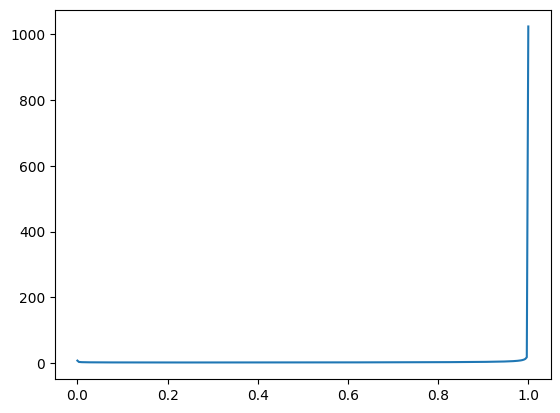}
    \caption{Diffusion-LM $g(t)^2$ for $t \in [0,1]$}
    \label{g2plot}
\end{figure}

Since the forward process is quite sharp, it has high second derivatives at the endpoints, as can be seen in \autoref{g2plot}. Training with the standard ELBO is unstable in this formulation. Thus, we train with the rescaled $\bm{x}$-prediction version of the ELBO, similar to what is done in the original Diffusion-LM paper. 

\begin{equation} \label{x prediction loss}
    \mathcal{L}_x = \mathbb{E}_{u(t) \, q(\bm{x}) \, q_{\varphi}(\bm{z}_t \mid \bm{x})}\left[\left\| E_\varphi({\bm{x}}) - \hat{E}_\varphi({\bm{z}_t,t})\right\|_2^2\right]
\end{equation}

\begin{figure}[H]
    \centering
    \begin{minipage}[b]{.45\textwidth}
        \includegraphics[width = 1\linewidth]{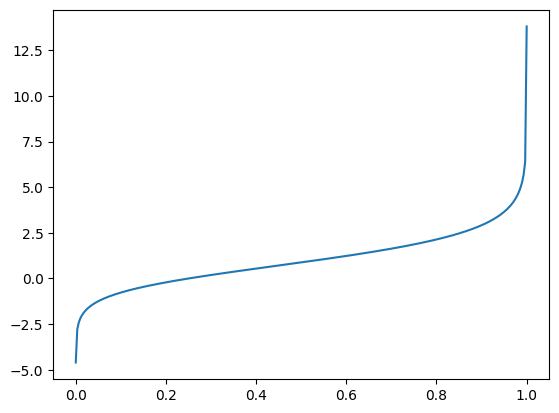}
        \caption{Diffusion-LM $\gamma(t)$ for $t \in [0,1]$}
        \label{Qstarch}
    \end{minipage}
    \begin{minipage}[b]{.45\textwidth}
        \includegraphics[width = 1\linewidth]{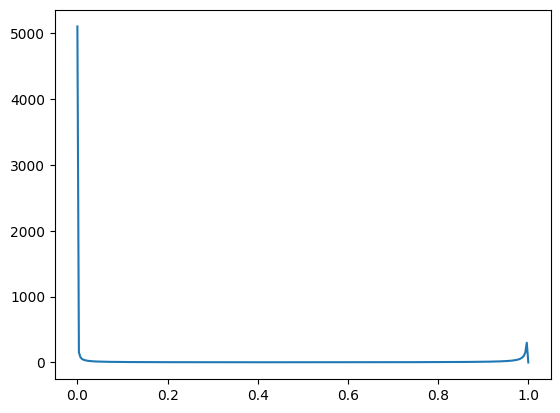}
        \caption{Diffusion-LM $\dot\gamma(t)$ for $t \in [0,1]$}
        \label{fig2}
    \end{minipage}
\end{figure}

\section{Sampling} \label{sec: sampling}

\begin{figure}[H]
    \centering

    \begin{minipage}[t]{.45\textwidth}
        \centering
        \begin{algorithm}[H]
            \caption{Sampling with an SDE}
            \label{continuous sampling}
            \begin{algorithmic}
                \Require $F_\varphi$, $g_\varphi$, $\bm{x}$, $T$, $E_\varphi$
                \State $\Delta t = 1/T$, $\textbf{z}_1 \sim p(\textbf{z}_1)$
                \For{ $t=1, \ldots, \frac{2}{T}, \frac{1}{T}$}
                    \State $\bar{\bm{w}} \sim \mathcal{N}(0,1)$
                    \State $\bm{z}_{t- \Delta t} = \bm{z}_t - f_\varphi\left(z_t, t, \hat{E}_\theta(\bm{z}_t, t)\right)\Delta t + g_\varphi(t) \bar{\bm{w}}\sqrt{\Delta t}  $
                \EndFor
                \State $\bm{x} \sim p_\varphi(\bm{x}|\bm{z}_0;E_\varphi)$
            \end{algorithmic}
        \end{algorithm}
        
    \end{minipage}%
    \hfill
    \begin{minipage}[t]{.45\textwidth}
        \centering
        \begin{algorithm}[H]
            \caption{Sampling with a conditional Markov chain}
            \label{discrete_sampling}
            \begin{algorithmic}
                \Require $F_\varphi$, $g_\varphi$, $\bm{x}$, $T$, $E_\varphi$, $\tilde{\sigma}_{s|t}$
                \State $\Delta t = 1/T$, $\textbf{z}_1 \sim p(\textbf{z}_1)$
                \For{ $t=1, \ldots, \frac{2}{T}, \frac{1}{T}$}
                    \State $s = t - \Delta t$
                    \State $\epsilon_{new} \sim \mathcal{N}(0,1)$
                    \State $\epsilon_{old} = F_{\varphi}^{-1}(\bm{z}_t, t, E_\varphi({\bm{x}})))$
                    \State $\tilde{\epsilon}_{s|t} = \sqrt{1-\tilde{\sigma}_{s|t}}\epsilon_{old}$  + $\sqrt{\tilde{\sigma}_{s|t}}\epsilon_{new}$
                    \State $\bm{z}_s = F_\varphi(\tilde{\epsilon}_{s|t}, s, \hat{E}_\theta(\bm{z}_t, t))$
                \EndFor
                \State $\bm{x} \sim p_\varphi(\bm{x}|\bm{z}_0;E_\varphi)$
            \end{algorithmic}
        \end{algorithm}
    \end{minipage}
\end{figure}

In the main text, we introduced two sampling approaches. However, the discrete sampling method discussed there represents only one possible way to sample from a discrete process in NFDM. Moreover, the described SDE can itself be converted into an ODE. This section explores additional sampling possibilities. First, we discuss sampling from the reverse SDE, explaining how it can be reformulated as an ODE at sampling time. Second, we note that sampling through the generative Markov chain is connected to the generative SDE via its marginals and asymptotic behavior as the number of steps approaches infinity. We then use these connections as a heuristic for sampling from the discrete process.

\subsection{Generative SDE} \label{sec: sampling sde}
To sample a sequence from NFDM, we can follow the sampling setup described in \cite{bartosh_neural_flow_2024}. Here we sample from the prior distribution $\bm{z}_1 \sim p(\bm{z}_1)$, and simulate the reverse SDE in \autoref{reversesde}. 

\begin{align} \label{generative sde}
    \mathrm{d}z_t &= \left[ f_\varphi\left(z_t, t, \hat{E}_\theta(\bm{z}_t, t)\right) 
    - \frac{g^2_\varphi(t)}{2} \nabla_{z_t} \log q_\varphi\left(z_t \mid \hat{E}_\theta(\bm{z}_t, t)\right) \right] \mathrm{d}t 
    + g_\varphi(t)\, \mathrm{d}\bar{w}.
\end{align}

\noindent We can vary $g_\varphi(t)$ during inference to control the level of stochasticity in the process. In the extreme case where $g_\varphi(t) \equiv 0$ \autoref{generative sde} becomes a deterministic ODE with the same marginals $q(\bm{z}_t)$ as the SDE. Deterministic sampling may increase the sampling speed compared to the stochastic method. 

\begin{align} \label{generative sde}
    \mathrm{d}\bm{z}_t &=  f_\varphi(z_t, t, \hat{E}_\theta(\bm{z}_t, t))
\end{align}

\noindent We can simulate the generative SDE using Euler-Maruyama discretization \cite{maruyama_continuous_1955}, and the generative ODE using the standard Euler method. The complete sampling procedure corresponding to these methods is shown in Algorithm \ref{continuous sampling}. Alternatively, adaptive ODE or SDE solvers can be employed to trade off between the number of function evaluations and solution accuracy. 

\subsection{Generative conditional Markov chain} \label{sec:sampling markov chain}
In \autoref{App: diff-LM}, we discuss how to convert from a discrete-time model to a continuous-time model with the same forward process. Similarly, we can convert from continuous to discrete. Here, we discuss this conversion and what might happen when we use a model trained with the continuous time loss to sample from the discrete-time process for a finite number of steps. The discrete-time process is specified using the forward conditional Markov chain,
\begin{equation}
    \bar{q}_{\varphi}(\bm{x}, \bm{z}_{0:1}) = q(\bm{x})q_{\varphi}(\bm{z}_0|\bm{x}) \left[ \prod_{t}\bar{q}_{\varphi}(\bm{z}_{t+\Delta t}|\bm{z}_t, \bm{x}) \right].
\end{equation}
which we approximate using the reverse conditional Markov chain,
\begin{equation} \label{reverse discrete}
    \bar{p}(\bm{x}, \bm{z}_{0:1}) = p(\bm{z}_1) \left[ \prod_{t}\bar{p}_{\theta}(\bm{z}_{t-\Delta t}|\bm{z}_t, \hat{E}_\theta(\bm{z}_t, t)) \right]p_{\theta}(\bm{x}|\bm{z}_0).
\end{equation}

\noindent We define the posterior distribution $\bar{q}(\bm{z}_{t+\Delta t}|\bm{z}_t, \bm{x})$ by describing how to sample $\bm{z}_s$ given $\bm{z}_t$ and $\bm{x}$ where $s < t$. To do this while preserving the correct marginal distribution $q_\varphi(\bm{z}_s|\bm{x})$. We can combine the Gaussian noise terms $\epsilon_s$ and $\epsilon_t$ into a new noise term $\tilde{\epsilon}_{s|t}$. We fill this into $F^{\text{Gaussian}}_{\varphi}(\tilde{\epsilon}_{s|t}, t, \bm{x})$ to sample $\bm{z}_s$. Specifically, we have:

\begin{align}
    \tilde{\epsilon}_{s|t} &= \sqrt{1-\tilde{\sigma}_{s|t}}\epsilon_{old}  + \sqrt{\tilde{\sigma}_{s|t}}\epsilon_{new}, \quad \text{gives} \\
\end{align}
\begin{align}
    z_{s} &= F^{\text{Gaussian}}_{\varphi}(\tilde{\epsilon}_{s|t}, t, E_\varphi({\bm{x}})) \\
    &= \mu_{\varphi} +  \sqrt{1-\tilde{\sigma}_{s|t}^2}\epsilon_{t} \sigma_{\varphi,s}  + \tilde{\sigma}_{s|t}\epsilon_{s}\sigma_{\varphi,s} \\
        &= \mu_{\varphi} +  \sqrt{\sigma_{\varphi,s}^2-\sigma_{\varphi,s}^2\tilde{\sigma}_{s|t}^2}\epsilon_{t}  + \tilde{\sigma}_{s|t}\sigma_{\varphi,s}\epsilon_{s} \label{this equation where we sample}
\end{align}

\noindent This result matches \cite{bartosh_neural_2024} in the fact that by marginalizing $\epsilon_t$ and $\tilde{\epsilon}_{s|t}$, we obtain a normal distribution with mean $\mu_\varepsilon$ and variance $\sigma_s^2 - \sigma_s^2 \tilde{\sigma}_{s \mid t}^2 + \sigma_s^2 \tilde{\sigma}_{s \mid t}^2 = \sigma_s^2$. Therefore, this sampling procedure satisfies $q_\varphi(\bm{z}_s \mid \bm{x}) = \int q_\varphi(\bm{z}_t \mid \bm{x}) \, \bar{q}_\varphi(\bm{z}_s \mid \bm{z}_t, \bm{x}) \, d\bm{z}_t$. Thus, the discrete forward process has the same marginals as the continuous process, regardless of the value we choose for $\tilde{\sigma}_{s|t}$. If we want the discrete forward process to match the continuous forward as $T\rightarrow\infty$, we can set $\tilde{\sigma}_{s|t}$ equal to an SNR-based quantity \cite{kingma_variational_2023}:

\begin{equation}
    \tilde{\sigma}_{s|t} = \tilde{\sigma}_{s|t}^* = 1 -\frac{SNR(t)}{SNR(s)}
\end{equation}

\noindent Note that this only works if we have easy access to the SNR, which is the case for Diffusion-LM, MuLAN. But for the more general forward process in NFDM, it is not. We can sample from the corresponding reverse process in \autoref{reverse discrete} by predicting $\hat{E}_\theta(\bm{z}_t, t)$ and filling this into \autoref{this equation where we sample}, the whole procedure for sampling is shown Algorithm \ref{discrete_sampling}.

There are two important observations about this sampling procedure. First, the equality between discrete and continuous only exists when we have an infinite number of timesteps. When we do not, bias will be injected into the sampling procedure. Second, the sampling procedure used in Algorithm \ref{discrete_sampling} approximates the posterior over embeddings given $\bm{z}_t$ by selecting a single prediction rather than sampling from the full conditional distribution. This deterministic approach reduces the variance of the sampling process and introduces a bias. The model is more likely to generate samples corresponding to the modes of the data distribution $q(\bm{x})$.

Since in some instances analytically determining $\tilde{\sigma}_{s|t}$ is not possible, we propose employing a further heuristic. We can treat $\tilde{\sigma}_{s|t} \in [0,1]$ as a tunable hyperparameter of the sampling process, and at test time select the samples that look best. Since tuning $\tilde{\sigma}_{s|t}$ may again inject bias, we should pay attention to metrics of distributional coverage as well as purely qualitative measures.
In the extreme case, when we choose to set $\sigma_{s|t} = 1$, we remove any dependence of the posterior of $\bm{z}_s$ on the previous sample $\bm{z}_t$. Thus, we sample from a series of marginal distributions as is written here:

\begin{equation} \label{reverse discrete star}
    \bar{p}(\bm{x}, \bm{z}_{0:1}) = p(\bm{z}_1) \left[ \prod_{t}\bar{p}_{\theta}(\bm{z}_{t-\Delta t}|\hat{E}_\theta(\bm{z}_t, t)) \right]p_{\theta}(\bm{x}|\bm{z}_0).
\end{equation}

This process is called a star diffusion because each $\bm{z}$ depends on the predicted datapoint, forming a star pattern. This method corresponds to the DDIM-style sampling method we presented in the main text. Star sampling may make the problem of sampling from only the modes of $q(\bm{x})$ described above more pronounced, since the intermediate latent variables now depend only on the collapsed posterior over the embeddings. Additionally, not taking into account any of the direction of the previous sample should make sampling more difficult and potentially introduce bias. 

In the discrete-time setting, adaptive timestep solvers are not available as in continuous-time SDEs. Instead, we can approximate adaptivity by using the learned importance sampling weights over $t$, drawing more samples from regions where the model experienced higher loss during training. This focuses sampling on more challenging regions, potentially improving efficiency and accuracy.

\section{Implementation Details} \label{implementation details}

We train three model variants: Diffusion-LM, MuLAN, and NFDM. This section describes the used architectures and hyperparameter. 

\subsection{Architecture}
For the predictor $\hat{E}_\theta(\bm{z}_t, t)$, we follow the parameterization of \citep{li_diffusion-lm_2022}, using BERT-base \citep{devlin_bert_2019} as the backbone with a dropout rate of 0.1 (approximately 90M parameters). The timestep $t$ is encoded using Fourier embeddings \citep{peebles_scalable_2023} passed through a two-layer MLP with SiLU activations \citep{elfwing_sigmoid-weighted_2017}. The resulting time embeddings are added to the token embeddings before being passed to the BERT encoder. We employ a word-level tokenizer, yielding a vocabulary of 10{,}767 entries.

MuLAN extends this setup by introducing an auxiliary latent variable $\bm{c}$ that conditions both the predictor and the forward process. The context $\bm{c}$ is encoded using a smaller BERT model (approximately 10M parameters) with hidden dimension equal to the embedding size $H$. This encoder is not time-conditioned. The predictor incorporates $\bm{c}$ by appending it as an additional token, while the forward process employs $\bm{\gamma}_{\varphi}(t, \bm{c})$ as in \citep{sahoo_diffusion_2024}, where polynomial coefficients are predicted by an MLP applied to $\bm{c}$.

NFDM uses two transformer backbones. The predictor reuses the same BERT-base model as above, while the forward process, which outputs $\overline{\mu}_{\varphi}(E_{\varphi}(\bm{x}), t)$ and $\overline{\sigma}_{\varphi}(E_{\varphi}(\bm{x}), t)$, uses a smaller BERT encoder (approximately 10M parameters). Because this model does not directly observe $\bm{z}_t$, it may struggle to condition on $t$. To address this, NFDM employs Adaptive LayerNorm conditioning \citep{peebles_scalable_2023, perez_film_2017} within the transformer blocks and before the final MLP.

\subsection{Training Setup}
All models use a batch size of 512, a hidden dimension of $H = 128$, and a sequence length of $S = 64$. Training runs for 800{,}000 iterations on NVIDIA A100 or H100 GPUs. Diffusion-LM and MuLAN require approximately 70--90 hours on an A100, while NFDM takes roughly 120 hours on an H100 due to the lack of flash attention support in the reverse process, the Jacobian Vector Product (JVP) computation for $\dot{F}_\varphi$ is incompatible with standard flash attention implementations. Overall, the total compute budget amounts to about 1{,}000 H100 hours and 2{,}000 A100 hours.

Diffusion-LM and MuLAN are trained using the Adam optimizer \citep{kingma_adam_2017} with a learning rate linearly decayed from $1\times10^{-4}$ and no weight decay. Gradients are clipped to a norm of 1.5 after 10K steps for the rescaled variants. MuLAN uses $\gamma_{\min} = -10$ and $\gamma_{\max} = 10$, while its rescaled version uses $\gamma_{\min} = -4.6$ and $\gamma_{\max} = 13.8$ to match the Diffusion-LM reference process.

NFDM is optimized jointly with Adam and Muon \citep{jordan2024muon}. Muon is applied to all multi-dimensional weight matrices within the transformer backbones, using a learning rate of 0.002 and momentum of 0.95, with gradient clipping to 0.3 after 10K steps. Adam is applied to all remaining parameters using the same configuration as for the other models.

\section{Additional results} \label{additional results}
In this section we provide further results complimenting those in \autoref{maintextresults}. We specifically look at modeling ablations for NFDM and MuLAN, and different choices for the sampling procedure. 
\subsection{Modeling ablations}
\begin{figure}[H]
  \centering

  \begin{minipage}[b]{.49\textwidth}
    \centering
    \includegraphics[width=\linewidth]{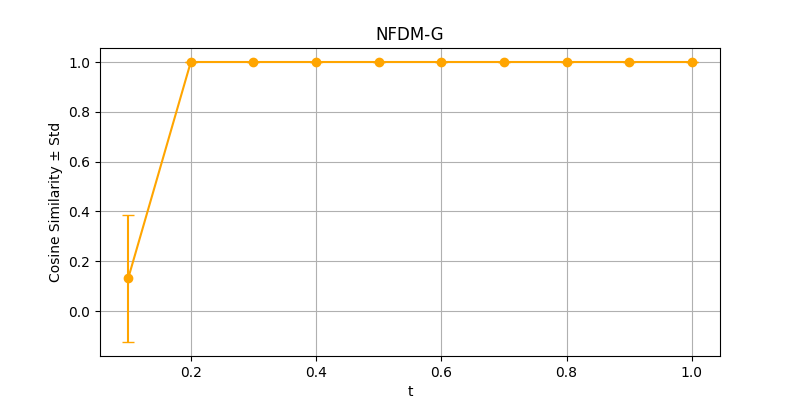}
  \end{minipage}
  \begin{minipage}[b]{.49\textwidth}
    \centering
    \includegraphics[width=\linewidth]{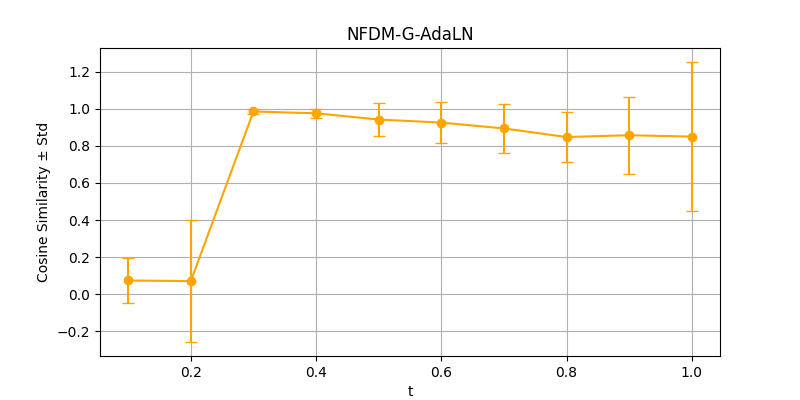}
  \end{minipage}

  \caption{Cosine‐similarity between $\overline{\mu}_{\varphi}(E_\varphi({\bm{x}}), t)$ and $\overline{\mu}_{\varphi}(E_\varphi({\bm{x}}), t+\Delta t)$, with steps of $t=0.1$ comparing NFDM-Additive (left) and NFDM (right). Standard deviation computed within batch.}
  \label{fig:both_models}
\end{figure}

We investigate whether the specific implementation of time conditioning in the forward process impacts textual quality. We compare NFDM with Adaptive Layernorm conditioning as presented in the main text to NFDM with time conditioning through additive timestep embeddings (NFDM-Additive). \autoref{fig:both_models} shows the cosine similarity for subsequent transformation of the input in the forward process for $t=\{0.1,0.2,0.3,0.4,0.5,0.6,0.7,0.8,0.9,1.0\}$. For NFDM-Additive, we observe that the model learns an input-dependent transformation of the data at $t=0.1$. However, subsequent transformations do not change with $t$. NFDM, by contrast, does learn a time-dependent transformation of the input, as shown on the right in \autoref{fig:both_models}.
\autoref{nfdm vs nfdm-AdaLN} shows that this results in a large improvement in perplexity for NFDM, although the model performs worse in terms of MAUVE, Diversity, and Memorization. Since NFDM achieves a perplexity similar to that of the data set and does not ignore $t$, we use this variant of the model for further experiments.

\begin{table}[H]
\centering
\caption{Quality of diffusion models with different time conditioning architectures.}
\begin{tabular}{lccccc}
\toprule
Conditioning & PPL $\downarrow$ & MAUVE $\uparrow$ & Diversity $\uparrow$ & Memorization $\downarrow$ \\
\midrule
Timestep Embeddings & 44.687 (0.731) & 0.688 (0.039) & 0.234 (0.002) & 0.100 (0.001) \\
Adaptive LayerNorm & 26.435 (0.221) &  0.426 (0.032) & 0.109 (0.001) & 0.143 (0.003) \\
\midrule
Dataset & 26.106  & 0.957 & 0.200 & 0.141 \\
\bottomrule
\end{tabular}
\label{nfdm vs nfdm-AdaLN}
\end{table}

\newpage
\subsection{Sampling ablations}

\subsubsection{Noise injection: generative conditional Markov chain}\label{markovchainres}
We begin with $\tilde{\sigma}_{s|t}$. As noted in \autoref{sec:sampling markov chain}, setting $\tilde{\sigma}_{s|t} = \tilde{\sigma}_{s|t}^*$ ensures asymptotic equivalence with the continuous-time formulation. Alternatively, setting $\tilde{\sigma}_{s|t} \in [0,1]$ can be used as a heuristic that retains the same marginals $q_{\varphi}(\bm{z}_t|\bm{x})$. We experiment with $\tilde{\sigma}_{s|t} \in \{1, 0.8, 0.5,\tilde{\sigma}_{s|t}^*\}$. Here, setting $\tilde{\sigma}_{s|t}=1$ corresponds to the DDIM-style sampling procedure used in the main text. For MuLAN-Rescaled, $\tilde{\sigma}_{s|t}^*$ is set according to the global SNR, and thus identical to the one used in Diffusion-LM. 

\autoref{big sigmast table} shows the results. Across all models, perplexity increases as $\tilde{\sigma}_{s|t}$ rises from 0.5 to 1. For models with a learned forward process, this increase in perplexity is accompanied by reduced diversity and memorization. This suggests that higher $\tilde{\sigma}_{s|t}$ values, which bring the process closer to star sampling, increase bias in the sampling trajectory, causing the models to sample more from high-probability modes. Supporting this, at $\tilde{\sigma}_{s|t} = 1$, all models achieve perplexities below the dataset baseline, indicating overly predictable sequences. 

We do not observe a consistent trend in MAUVE. The reason for this could be that MAUVE is based on GPT-2 Large sequence encodings, which it uses to compute distributional coverage. Thus, it benefits from both improved predictability under GPT-2 (PPL) and improved diversity. This leads to a trade-off when the two move in opposite directions.

Comparing fixed $\tilde{\sigma}_{s|t}$ values to the asymptotically correct $\tilde{\sigma}_{s|t}^*$, we find that the latter results in higher diversity and lower memorization, but also higher perplexity and lower MAUVE. This indicates that while the sampling trajectory may be less biased under $\sigma_{s|t}^*$, this leads to a considerable loss in sample quality.

\begin{table}[H]
\centering
\caption{Quality of diffusion model samples when varying schedule of $\tilde{\sigma}_{s|t}$}
\begin{NiceTabular}{lccccc}[cell-space-top-limit=3pt,cell-space-bottom-limit=3pt]
  \CodeBefore
    \rowcolor{gray!25}{6-9}  
  \Body
  \toprule
  Method                        & $\sigma_{s|t}$   &  PPL $\downarrow$ & MAUVE $\uparrow$ & Diversity $\uparrow$ & Memorization $\downarrow$ \\
  \midrule
  \multirow{4}{*}{Diffusion-LM} 
                                & 1.0               & 21.05 (0.42) & 0.72 (0.04) & 0.18 (0.00) & 0.14 (0.00) \\
                                & 0.8               & 21.05 (0.24) & 0.76 (0.01) & 0.18 (0.00) & 0.14 (0.00) \\
                                & 0.5               & 21.68 (0.22) & 0.74 (0.02) & 0.18 (0.00) & 0.14 (0.00) \\
                                & $\tilde{\sigma}_{s|t}^*$  & 64.47 (0.90) & 0.68 (0.06) & 0.28 (0.00) & 0.09 (0.00) \\
  \multirow{4}{*}{MuLaN-Rescaled} 
                                & 1.0               & 23.10 (0.20) & 0.60 (0.04) & 0.18 (0.00) & 0.12 (0.00) \\
                                & 0.8               & 27.71 (0.31) & 0.60 (0.03) & 0.20 (0.00) & 0.12 (0.00) \\
                                & 0.5               & 34.02 (0.26) & 0.62 (0.06) & 0.22 (0.00) & 0.11 (0.00) \\
                                & $\tilde{\sigma}_{s|t}^*$  & 124.40 (1.60) & 0.18 (0.01) & 0.36 (0.00) & 0.06 (0.00) \\
   \multirow{4}{*}{NFDM} 
                                & 1.0               &  23.52 (0.36) & 0.42 (0.02) & 0.10 (0.00) & 0.15 (0.00) \\
                                & 0.8               &  26.44 (0.22) & 0.43 (0.03) & 0.11 (0.00) & 0.14 (0.00) \\
                                & 0.5               &  30.23 (0.22) & 0.50 (0.06) & 0.13 (0.00) & 0.13 (0.00) \\ 
                                & $\tilde{\sigma}_{s|t}^*$  &  - & - & - & -   \\    
  \bottomrule
\end{NiceTabular}
\label{big sigmast table}
\end{table}

\subsubsection{Noise injection: generative SDE}\label{generativesderes}
We now turn to continuous sampling, examining both the stochastic (SDE) and deterministic (ODE) variants, and comparing their performance to discrete-time sampling. These settings are governed by the volatility schedule $g^2(t)$, as discussed in \autoref{sec: sampling sde}.

\autoref{SDE ODE table } presents the results. Broadly, sampling from the continuous-time formulation results in worse perplexity compared to discrete sampling. However, NFDM breaks this trend: when sampled via the SDE, it performs on par with or better than Diffusion-LM under $\tilde{\sigma}_{s|t}^*$, achieving similar PPL, higher MAUVE, increased diversity, and lower memorization. Compared to using discrete sampling for NFDM, we see that the generated samples have higher diversity, higher mauve, and lower memorization. This suggests that using the generative SDE can reduce sampling bias, leading to broader distributional coverage and less mode-seeking behavior.

Comparing the SDE to the ODE directly, we find that SDE sampling consistently outperforms its deterministic counterpart for almost all models. The exception is Diffusion-LM. In this case, SDE sampling leads to significantly degraded performance. We attribute this to the design of its volatility schedule $g^2(t)$, which exhibits large spikes near the endpoints ($t=0$ and $t=1$), injecting excessive noise late in the trajectory, which leads to poor generation quality. 

\begin{table}[H]
\centering
\caption{Quality of diffusion model samples when sampled with generative SDE versus generative ODE}
\begin{NiceTabular}{lccccc}[cell-space-top-limit=3pt,cell-space-bottom-limit=3pt]
  \CodeBefore
    \rowcolor{gray!25}{4-5}  
  \Body
  \toprule
  Method                        & $g_\varphi(t)$   &  PPL $\downarrow$ & MAUVE $\uparrow$ & Diversity $\uparrow$ & Memorization $\downarrow$ \\
  \midrule
  \multirow{2}{*}{Diffusion-LM} 
                                & SDE   & 2658.17 (21.67) & 0.01 (0.00) & 0.78 (0.00) & 0.00 (0.00) \\
                                & ODE   & 74.07 (1.92) & 0.42 (0.02) & 0.29 (0.01) & 0.09 (0.00) \\
  \multirow{2}{*}{MuLaN-Rescaled} 
                                & SDE   &  116.61 (3.37) & 0.20 (0.03) & 0.35 (0.01) & 0.06 (0.00) \\ 
                                & ODE   &  145.54 (1.49) & 0.09 (0.02) & 0.36 (0.00) & 0.06 (0.00) \\
   \multirow{2}{*}{NFDM} 
                                & SDE   & 68.33 (1.45) & 0.76 (0.05) & 0.32 (0.00) & 0.07 (0.00) \\
                                & ODE   & 289.14 (6.14) & 0.08 (0.02) & 0.47 (0.00) & 0.05 (0.00) \\
  
  \bottomrule
\end{NiceTabular}
\label{SDE ODE table }
\end{table}

\subsubsection{Sampling steps} \label{samplingstepsres}
In this section, we evaluate the impact of reducing the number of sampling steps to values equal to or below the sequence length. This setting gives the model at most one step per token. We hypothesize that models which learn the forward process, conditionally on time $t$ and/or data $\bm{x}$, will produce smoother sampling trajectories. Consequently, we expect them to perform better than models with fixed forward processes when operating on a sparse timestep grid.

\autoref{sigmast 64} presents the results for 64 and 32 sampling steps with discrete sampling and $\tilde{\sigma}_{s|t}=1$. We find that Diffusion-LM, which uses a static forward process, handles low-step generation relatively well. In contrast, NFDM struggles significantly, especially at 32 steps, where MAUVE drops to zero and perplexity increases sharply, indicating that many generated sequences become unintelligible. \cite{bartosh_neural_flow_2024} describes a way to straighten the learned trajectories with an additional loss term, which could significantly improve the few-step performance of the models with a learned forward process. We defer this to future work. \autoref{SDE 64} shows that the results for the sampling with generative SDE are much worse than for the discrete procedure, across all models. 

These results do not align with the findings of \cite{bartosh_neural_flow_2024}, which report that NFDM models outperform fixed-forward process baselines under few-step sampling. 

\begin{table}[H]
\centering
\caption{Quality of diffusion model samples when decreasing the number of sampling steps (NFE). Samples are obtained using the generative Markov chain}
\begin{NiceTabular}{lccccc}[cell-space-top-limit=3pt,cell-space-bottom-limit=3pt]
  \CodeBefore
    \rowcolor{gray!25}{4-5}  
  \Body
  \toprule
  Method                        & NFE   &  PPL $\downarrow$ & MAUVE $\uparrow$ & Diversity $\uparrow$ & Memorization $\downarrow$ \\
  \midrule
  \multirow{2}{*}{Diffusion-LM} 
                                & 64   &  29.55 (0.29) & 0.81 (0.01) & 0.21 (0.00) & 0.13 (0.00) \\
                                & 32   &  38.23 (0.81) & 0.77 (0.03) & 0.22 (0.00) & 0.12 (0.00) \\
  \multirow{2}{*}{MuLaN-Rescaled} 
                                & 64   & 47.68 (0.63) & 0.57 (0.04) & 0.26 (0.00) & 0.09 (0.00) \\
                                & 32   & 72.97 (0.82) & 0.23 (0.03) & 0.28 (0.00) & 0.08 (0.00) \\
   \multirow{2}{*}{NFDM} 
                                & 64   & 141.87 (2.76) & 0.19 (0.01) & 0.39 (0.00) & 0.05 (0.01) \\
                                & 32   & 571.78 (10.90) & 0.02 (0.00) & 0.41 (0.01) & 0.02 (0.00) \\
  
  \bottomrule
\end{NiceTabular}
\label{SDE 64}
\end{table}

\begin{table}[H]
\centering
\caption{Quality of diffusion model samples when decreasing the number of sampling steps (NFE). Samples are obtained using the generative SDE}
\begin{NiceTabular}{lccccc}[cell-space-top-limit=3pt,cell-space-bottom-limit=3pt]
  \CodeBefore
    \rowcolor{gray!25}{4-5}  
  \Body
  \toprule
  Method                        & NFE   &  PPL $\downarrow$ & MAUVE $\uparrow$ & Diversity $\uparrow$ & Memorization $\downarrow$ \\
  \midrule
  \multirow{2}{*}{Diffusion-LM} 
                                & 64   & 3950.27 (24.05) & 0.00 (0.00) & 0.58 (0.00) & 0.00 (0.00) \\
                                & 32   & 3707.34 (55.64) & 0.00 (0.00) & 0.55 (0.01) & 0.00 (0.00) \\
  \multirow{2}{*}{MuLaN-Rescaled} 
                                & 64   & 156.06 (1.15) & 0.10 (0.01) & 0.38 (0.00) & 0.05 (0.00) \\
                                & 32   & 205.21 (2.77) & 0.04 (0.00) & 0.41 (0.00) & 0.04 (0.00) \\
   \multirow{2}{*}{NFDM} 
                                & 64   & 7181.29 (167.44) & 0.00 (0.00) & 0.90 (0.00) & 0.00 (0.00) \\ 
                                & 32   & 363.41 (14.76) & 0.00 (0.00) & 0.38 (0.01) & 0.00 (0.00) \\
  
  \bottomrule
\end{NiceTabular}
\label{sigmast 64}
\end{table}

\newpage
\subsection{Generated samples} \label{gensamplesdiscussion}
Below, we show two generated samples for each of the three best-performing diffusion models. We use discrete sampling with $\tilde{\sigma}_{s|t}=1$, for which we achieved the lowest perplexity. The samples displayed here have a whitespace between each of the generated tokens. To evaluate the samples correctly, we post-process them with a simple function that removes trailing whitespace and whitespace between tokens that should be one word.

\begin{figure}[H] \label{goodsamples}
  \begin{subfigure}[t]{\textwidth}
    \caption{\textbf{Diffusion-LM}}
    \begin{quote}
      START My friends and I went for a hike in the woods . The weather was very crowded when we got there . When we got there it started to rain . We sat for so long until it started to pour . We had to go back to our car to go out . END PAD PAD PAD PAD PAD PAD PAD PAD"
    \end{quote}
    \begin{quote}
     START Kelly really wanted a new haircut . Luckily she could n't find one she wanted . Thankfully she ordered one . When she got it on it looked great . Kelly was thrilled . END PAD PAD PAD PAD PAD PAD PAD PAD PAD PAD PAD PAD PAD PAD PAD PAD PAD PAD PAD PAD PAD PAD PAD PAD PAD PAD PAD PAD
    \end{quote}
  \end{subfigure}
  \begin{subfigure}[t]{\textwidth}
    \caption{\textbf{MuLAN-Rescaled}}
    \begin{quote}
      START Kate bought a pair of pretty boots . They were longer than she expected . Because of them were too small for them . Kate had to wear them out the next day . Kate wore them to school again . END PAD PAD PAD PAD PAD PAD PAD PAD PAD PAD PAD PAD PAD PAD PAD PAD PAD PAD PAD PAD PAD
    \end{quote}
    \begin{quote}
    START Last night I made pasta . I used a new recipe . It tasted amazing . It still tasted amazing . I wanted to try it . END PAD PAD PAD PAD PAD PAD PAD PAD PAD PAD PAD PAD PAD PAD PAD PAD PAD PAD PAD PAD PAD PAD PAD PAD PAD PAD PAD PAD PAD PAD PAD PAD PAD PAD PAD
    \end{quote}
  \end{subfigure}

  \begin{subfigure}[t]{\textwidth}
    \caption{\textbf{NFDM}}
    \begin{quote}
      START Tom needed to get a pair for school . He realized he had to get a new pair of shoes . He realized that the store was too high . He went and saw his shoes were in stock . He had to get a good deal for his shoes . END PAD PAD PAD PAD PAD PAD PAD PAD PAD PAD PAD
    \end{quote}
    \begin{quote}
      START Billy and his mother decided to go to the beach . They got in the car and drove to the beach . The UNK told them they would take this trip . They got to the beach and got a lot of sand . They finally got in the car and drove to the beach . END PAD PAD PAD PAD PAD PAD
    \end{quote}
  \end{subfigure}

\caption{Samples generated using discrete sampling with $\tilde{\sigma}_{s|t}=1$ and 2000 steps.}
\label{goodsamples}
\end{figure}

\newpage
\begin{figure}[H]
  \begin{subfigure}[t]{\textwidth}
    \caption{\textbf{Diffusion-LM}}
    \begin{quote}
      START Jim was on a road trip . He had been driving for a while when the bell rang . He had never been to a hotel . He got into the hotel and a hotel when he got to . He had a great time . END PAD PAD PAD PAD PAD PAD PAD PAD PAD PAD PAD PAD PAD PAD PAD PAD
    \end{quote}
    \begin{quote}
      START John did n't have anything to do with his family . He decided that he should stay home to work alone . He called his boss and offered to help . John accepted the offer . He now has more money at home with his family . END PAD PAD PAD PAD PAD PAD PAD PAD PAD PAD PAD PAD PAD PAD PAD
    \end{quote}
  \end{subfigure}
  
  \begin{subfigure}[t]{\textwidth}
    \caption{\textbf{MuLAN-Rescaled}}
    \begin{quote}
      START UNK had been craving pizza lately . She asked her husband if they could eat it . He gave her a box of stale pizza . He told her they were sweet . UNK now eats at fifteen times . END PAD PAD PAD PAD PAD PAD PAD PAD PAD PAD PAD PAD PAD PAD PAD PAD PAD PAD PAD PAD PAD PAD
    \end{quote}
    \begin{quote}
      START I was getting a divorce . My wife wanted me out of her closet every single . It was cold and UNK . I tried to dig on them . My wife . me on a hand and got rid of ! END PAD PAD PAD PAD PAD PAD PAD PAD PAD PAD PAD PAD PAD PAD PAD PAD PAD PAD PAD PAD
    \end{quote}
  \end{subfigure}
  \begin{subfigure}[t]{\textwidth}
    \caption{\textbf{NFDM}}
    \begin{quote}
      START Rob was wondering Rob Rob could tell Rob was then Rob proposed to race but was determined to rob Rob Rob He decided to give up Rob UNK Rob Rob Rob Finally Rob and afterwards END PAD PAD PAD PAD PAD PAD PAD PAD PAD PAD PAD PAD PAD PAD PAD PAD PAD PAD PAD PAD PAD PAD PAD PAD PAD PAD PAD
    \end{quote}
    \begin{quote}
      START The kids were traveling through China . They got bored until it opened and it was beautiful ! They spent hours on making their own favorite meal . They made so much food ! END PAD PAD PAD PAD PAD PAD PAD PAD PAD PAD PAD PAD PAD PAD PAD PAD PAD PAD PAD PAD PAD PAD PAD PAD PAD PAD PAD PAD
    \end{quote}
  \end{subfigure}

\caption{Samples generated using discrete sampling with $\tilde{\sigma}_{s|t}=1$ and 64 steps.}
\label{bad samples}
\end{figure}

\end{document}